\documentclass[
	a4paper, 
	10pt, 
	unnumberedsections, 
	twoside, 
]{LTJournalArticle}

\usepackage{newunicodechar}
\newunicodechar{ }{~} 

\usepackage{wrapfig}
\usepackage{amsmath}
\usepackage{float}
\usepackage{booktabs}
\usepackage{stfloats}
\usepackage{makecell}
\usepackage{booktabs}
\usepackage{multirow}
\usepackage{caption}
\usepackage[table]{xcolor}
\definecolor{sullycyan}{HTML}{79e2fc}

\usepackage{authblk}

\date{}  
\renewcommand{\date}[1]{}  

\usepackage[most]{tcolorbox}
\tcbuselibrary{breakable} 

\usepackage[font=small,skip=2pt]{caption}

\DeclareUnicodeCharacter{2500}{-}       
\DeclareUnicodeCharacter{251C}{\ensuremath{\mid}}  
\DeclareUnicodeCharacter{2514}{\ensuremath{\bot}}  
\DeclareUnicodeCharacter{2009}{\,}      

\title{Second Opinion Matters: \\ Towards Adaptive Clinical AI via The Consensus of Expert Model Ensemble} 

\author[1,2]{Amit Kumthekar}
\author[1]{Zion Tilley}
\author[1]{Henry Duong MD}
\author[1]{Bhargav Patel MD}
\author[1]{Michael Magnoli}
\author[1]{Ahmed Omar}
\author[1]{Ahmed Nasser}
\author[1]{Chaitanya Gharpure PhD}
\author[1]{Yevgen Reztzov}

\affil[1]{Sully AI}
\affil[2]{Stanford University}

\renewcommand{\maketitlehookd}{%
  \begin{center}
    \textbf{Abstract}
  \end{center}
  \vspace{-0.5em}
  \begin{center}
    \begin{minipage}{0.85\textwidth}  
      \small  
      \noindent
      
\textit{Despite the growing clinical adoption of large language models (LLMs), current approaches heavily rely on single-model architectures. To overcome risks of obsolescence and rigid dependence on single model systems, we present a novel  framework, termed the \textit{Consensus Mechanism}, which has been designed to flexibly integrate the strengths of multiple "expert'" models. Mimicking clinical triage and multidisciplinary clinical decision-making, the Consensus Mechanism implements an ensemble of specialized medical "expert" agents enabling unprecedented performance on medical benchmarks and robust adaptability as newer LLMs become available. This process enables the Consensus Mechanism to be optimized for cost, latency, or performance, purely based on the interior model configuration.
}
\vspace{1em}

\textit{To rigorously evaluate the Consensus Mechanism, we employed three medical evaluation benchmarks: MedMCQA, MedQA, and MedXpertQA Text, and  the differential diagnosis dataset, DDX+. When optimized for performance, our method surpasses state of the art (SoTA) reasoning models, such as OpenAI's O3 and Google's Gemini 2.5 Pro. On MedXpertQA, the Consensus Mechanism achieved an accuracy of  $61.0\%$ compared to $53.5 \%$ and $45.9\%$ for OpenAI's O3 and Google's Gemini 2.5 Pro. Improvement was consistent across benchmarks with an increase in accuracy on MedQA ($\Delta \text{Accuracy}_{\text{consensus-O3}} = 3.4\%$) and MedMCQA ($\Delta \text{Accuracy}_{\text{consensus-O3}} = 9.1\%$). These accuracy gains extended to differential diagnosis generation, where our system demonstrated improved recall and precision ($F1_{\text{consensus}} = 0.326$ vs. $F1_{\text{O3-high}} = 0.2886$) and a higher top-1 accuracy for DDx ($Top 1_{\text{ consensus}} = 52.0\%$ vs. $Top 1_{\text{ O3-high}} = 45.2\%$). Together, these results highlight the Consensus Mechanism as a robust and scalable approach for augmenting the clinical-decision making capabilities of publicly available large language models. 
}
    \end{minipage}
  \end{center}
}

\begin{document}

\maketitle 


\section{1 Introduction}
The past few years have seen an accelerated adoption of large language models and generative artificial intelligence into clinical practice\textsuperscript{\cite{1}}. However, the rapid evolution of these models presents a double-edged sword\textsuperscript{\cite{2},\cite{3}}. Although newer models often improve clinically relevant capabilities, the frequent emergence of novel models poses a challenge for sustainable single-model deployment in a healthcare setting\textsuperscript{\cite{2},\cite{4}}.\\

The pace of advancement in the field makes reliance on any single model increasingly impractical. High-performing models are frequently surpassed or updated, and shifts in proprietary access, API behavior, or cost structures make it difficult to maintain stable clinical workflows\textsuperscript{\cite{4},\cite{5}}. Furthermore, many leading models remain closed-source and cloud-dependent, introducing additional concerns around privacy and security, especially in regulated healthcare settings\textsuperscript{\cite{2},\cite{4},\cite{5}}.

Despite increasing medical benchmark scores and improved reasoning capabilities, current large language models (LLMs) continue to fall short in the execution of the nuanced, complex reasoning that is required for real world clinical care\textsuperscript{\cite{6},\cite{7},\cite{8}}. These limitations are particularly problematic in medicine, where subtle contextual cues, probabilistic thinking, and justified decision-making are essential to safe and effective care\textsuperscript{\cite{8}}. To measure these improvements in clinical reasoning, we rely on the MedXpertQA dataset, which was designed to evaluate performance on complex clinical  tasks\textsuperscript{\cite{9}}.

\noindent Taken together, these challenges reveal a critical gap in the clinical application of LLMs. In response, there is growing interest in modular, adaptive frameworks that can aggregate the strengths of multiple models while minimizing reliance on a single, large model. This approach draws inspiration from expert-based systems, which have proven to be successful across a wide range of domains\textsuperscript{\cite{10}}. An expert-based round-table-like approach provides a structured method for synthesizing diverse viewpoints, while achieving contextually nuanced output\textsuperscript{\cite{11}}. This inherently human behavior of conversation and deliberation can be applied to solving complex, real-world problems using LLMs.

\noindent In this paper, we introduce the Consensus Mechanism, or Sully Medical Consensus 1 (\textit{MedCon-1}), a dynamic and modular framework designed to meet the aforementioned challenges with current methods. Through the orchestration of multiple independently configurable LLMs as domain-specific "expert agents", the system simulates real-world clinical decision making via downstream expert deliberation\textsuperscript{\cite{12}}. 

As shown in Figure 1, our modular consensus system outperforms leading single-model systems across diverse multiple choice medical benchmarks, including MedQA, MedMCQA, and most notably, MedXpertQA Text.  These gains in accuracy highlight not only the promise of expert-agent deliberation, but also the potential clinical utility of a modular and adaptable LLM architecture.

\begin{figure}[!htbp]
\centering
\vspace{-.5em} 
\includegraphics[width=\columnwidth]{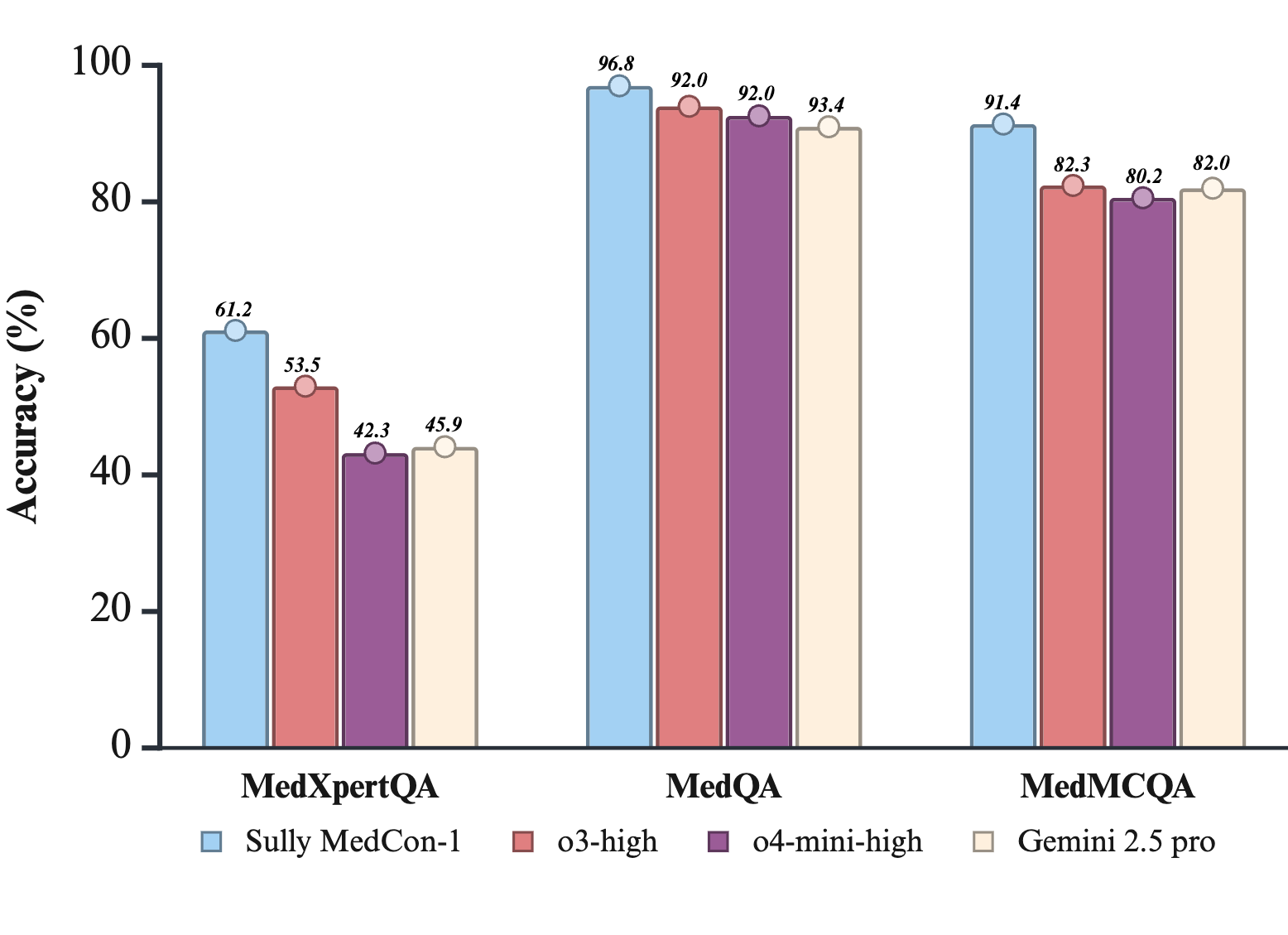}
\vspace{-1em} 
\caption{Consensus demonstrates increased accuracy across all evaluation benchmarks, with a notable increase in accuracy for \textit{MedXpertQA} Text.}
\label{fig:consensus-overview}
\vspace{-1em} 
\end{figure}


\section{2 Methodologies}
\subsection{2.1 Design Overview }

At its core, the Consensus Mechanism is a modular clinical reasoning framework that synthesizes the best aspects of each expert to generate a robust and diverse response for a given task. Unlike a traditional Chain-of-Thought (CoT) or standard Mixture-of-Experts (MoE) approach, the Consensus Mechanism builds on both of these methods by defining the role of each expert through a chain of thought like decomposition. This is accomplished by a triage model which assigns the relevant medical specialists based on a clinical task. Each expert then independently analyzes the task explicitly from its respective specialist perspective and reports back a probability distribution of correct answers. Embracing the interdisciplinary and probabilistic nature of medicine, the individual distributions are then aggregated and weighted. A final consensus model is then given each expert's analysis, and the recalculated probability distribution, allowing it to synthesize a unified, well-supported output. 

\begin{figure}[t]
  \centering
  \includegraphics[width=\columnwidth]{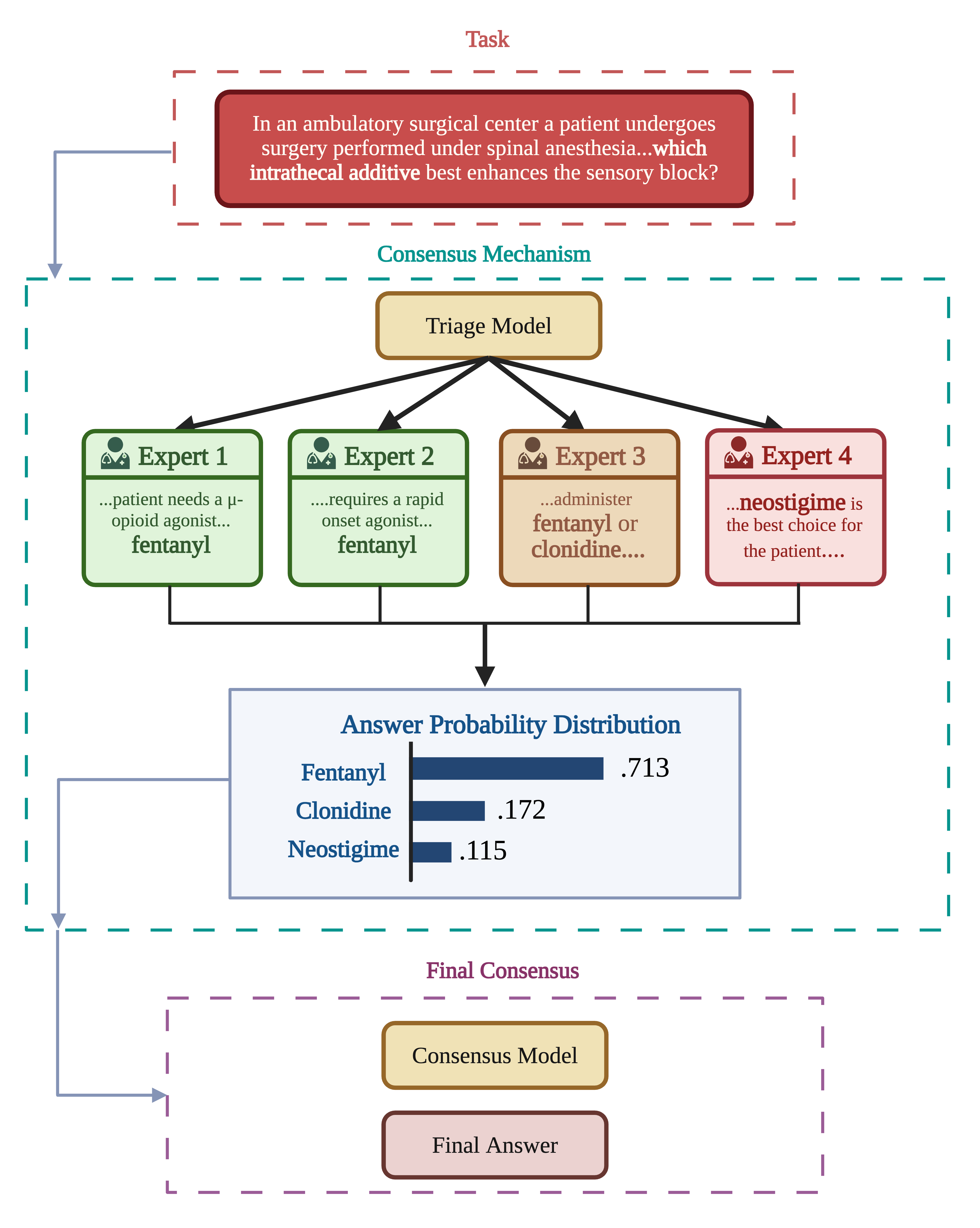} 
  \caption{An overview of the Consensus Mechanism's pipeline. A task is fed to a triage model which picks the composition of an expert block to support the consensus model. Context and a probability distribution from the experts are provided to a consensus model for final determination}
  \label{fig:consensus-overview}
\end{figure}

\subsection{2.2 Expert Based Decomposition }

Consensus implements a triage-inspired system that delegates a respective query to a panel of specialized LLM experts. Each expert is instantiated as a distinct, publicly available model configured to a particular medical specialty. This configuration is determined based on the task analysis outputted by the triage agent.

The triage agent orchestrates this initial decomposition by identifying the type of medical task and dividing the problem among a synergistic group of domain-specific models\textsuperscript{\cite{13}}. When prompted, the triage agent first performs a preliminary assessment to infer the task type, then routes the query to the most appropriate set of downstream medical specialists. The preliminary analysis enables the agent to draw out important contextual signals from the input query, improving the relevance and coverage of the selected experts.

We define the input medical query as Q, which is processed by a triage LLM to return the task type and best suited medical specialties to handle the respective query: 
\begin{align*}
  \mathrm{LLM}_{\mathrm{triage}}(Q) 
  &\;\Rightarrow\; (\text{Task Type},\,\text{Specialties})
  \\[0.5ex]
  &\;\Rightarrow\; \text{Specialties}
    = \{\,s_{1},\,s_{2},\dots,s_{n}\}.
\end{align*}

Each specialty, \(s_{i}\in\{s_{1},s_{2}\dots s_{n}\}\), is mapped to a corresponding expert model, with index i, which processes the query using domain-specific reasoning to generate its respective response \(R_{\mathrm{expert}}^{(i)}\):

\[
R_{\mathrm{expert}}^{(i)}
\;=\;
\mathrm{LLM}_{s_{i}}\bigl(\text{Query},\,\text{Task Type}\bigr).
\]

To compose the final output, \(O_{\mathrm{final}}\), we pass each expert's answer and probability distribution through a probability-reweighting function, \(f\), \textit{(see section 2.3-2.4}):

\[
  O_{\mathrm{final}}
  \;=\;
  f\bigl(R_{\mathrm{expert}}^{(1)}, \dots, R_{\mathrm{expert}}^{(n)}\bigr).
\]

This aggregated group of responses is then provided to a final consensus LLM which produces the overall answer:
\[
  A_{\mathrm{final}}
  \;=\;
  A_{\mathrm{consensus}}
  \;=\;
  \mathrm{LLM}_{\mathrm{consensus}}\bigl(O_{\mathrm{final}}\bigr).
\]\

The overall design builds on the fundamental principles of  CoT reasoning and MoE modeling. Our approach achieves a similar end goal of structured reasoning, but by distributing the reasoning across multiple domain-focused agents instead of a single model's internal sub-experts\textsuperscript{\cite{13},\cite{14}}.  Though MoE generally applies to a single-model system, we extrapolated the technique to a multi-model expert block to leverage increased diversity\textsuperscript{\cite{15}}. Each independent viewpoint captures the multi-dimensional nature of clinical reasoning without placing overwhelming pressure on any one model. This allows consensus to cover a broader diagnostic space than any single generalist model.


\subsection{2.3 Probability Weighting}

During the development of the Consensus Mechanism, we observed issues pertaining to overconfidence and  the inaccurate representation of potential answer choices. The key design choice when addressing this issue was to have each specialist LLM return not just a single answer, but a full probability distribution over the set of possible answers or diagnoses. Obtaining a full probability distribution effectively captures each expert's uncertainty and highlight subtle variations in their clinical judgment. This allowed the Consensus Mechanism to offer a nuanced and transparent overview of each expert's decision-making process. In clinical practice, understanding the confidence level behind a diagnosis can be just as critical as the diagnosis itself, ultimately supporting more informed and transparent clinical decision.

To utilize the probabilities generated by each expert, we combine them into a single aggregated distribution that reflects the initial consensus of the experts. We adopted a Weighted Log Opinion Pool (WLOP) approach because of its ability to adjust the contribution of different experts to the probability distribution (\(P_{\mathrm{combined}}(X)\)) \textsuperscript{\cite{16},\cite{17}}. This technique computes the weighted geometric mean of each expert's probability distribution, where \(p_{i}(X)\) is the probability assigned to the outcome \(X\) by expert \(i\). \

\(w_{j}\) is an adjustable parameter describing the contribution of expert \(j\) to the final distribution where:

\[
w_{i }>0,\;\sum_{}^{}w_{i}=1.
\]
To represent these probabilities in log‐space, we apply a log‐transformation and compute a weighted sum in log‐space for each outcome index \(i\):

\[
P_{\text{combined}}^{}(X) = \sum_{i} w_{i} \log p_{i}(X)
\]

Each \(P_{\text{combined}}^{}(X)\) is normalized, \(P_{\text{normalized}}^{}(X)\), using a softmax-like function across indices \(X^{j}\) of all answers:

\[
P_{\text{normalized}}(X) = 
\frac{e^{P_{\text{combined}}^{}(X)}}
     {\sum_{} e^{P_{\text{combined}}^{}(X^{j})}}
\]

This method provides a flexible and interpretable framework for integrating expert judgments, naturally rewarding confident and consistent agreement while moderating the influence of uncertain or divergent opinions. Conversely, if all experts are uncertain, the normalization process yields a more even distribution, appropriately reflecting the inherent ambiguity of the task. This prevents the consensus model from being inadvertently biased toward incorrect conclusions.

By merging the experts' outputs in this probabilistic manner, we retain the subtle information each model provides. Instead of a rudimentary vote, the consensus model considers how strongly each expert favored an answer allowing it to make a much more fine-grained and robust decision. Ultimately, this approach adds an important layer of quantitative confidence modeling to the system, aligning with the probabilistic reasoning approaches often used by clinicians.  Therefore, we anticipated this functionality to be especially useful for edge cases where decisions are nuanced and a definitive answer may be unclear.


\subsection{2.4  Cascade Boosting for Probabilities}

To better capture nuances in possible answers, our consensus system employs a cascade boosting mechanism, designed to enhance the probability of answers based on their frequency across experts' responses. Cascade boosting systematically considers multiple ranks to reinforce answers frequently appearing in subsequent positions, with the hope of reflecting broader expert uncertainty and improving robustness in the final consensus. This approach is particularly beneficial in ambiguous clinical scenarios, where a consensus might not be immediately apparent from top choices alone or for complex tasks where experts may pose multiple plausible answers.

Using the aggregated and weighted probability distribution  \textit{(see 2.3 Probability Weighting)} a frequency table is established based on how frequently each candidate answer appears at each ranked position. \( f_{X,r} \) represents the frequency of answer \( X \) appearing at rank \( r \). Calculating this rank frequency data highlights plausible answer alternatives, even if these answers are not always the primary choice.

The system then applies a cascade weighting scheme to these rank frequencies. Cascade weights are predetermined values that progressively diminish with each lower rank, reflecting the decreasing influence of answers appearing in lower-ranked positions. To preserve the probability of the most likely answer, the weight at rank one, \( \theta_1 \), was set to one with subsequent weights equal to half the previous ranks weight:
\begin{gather*}
\theta_{r+1} = \frac{1}{2} \theta_r \\
\theta_r = [1.0,\; 0.5,\; 0.25,\; 0.125,\; 0.0625,\; 0.03125]
\end{gather*}

\noindent
Higher-ranked positions thus have greater impact on the boosting process, accurately mirroring clinical judgment hierarchies. These values then factor into a probability boosting calculation which can be controlled by a boost scaler, \( \lambda_{\text{boost}} \), a tunable parameter that adjusts the relative influence of the overall probability boost. The boosted score for each answer is computed using the following equation:
\[\text{BoostedScore}(X) = P_{\text{normalized}}(X) + \lambda_{\text{boost}} \cdot \sum_{}^{} \left(f_{X,r} \times \theta_r \right)\]
\noindent
Following boosting, the final aggregated probability distribution (\(P^{\text{final}}(X)\)) is composed by re-normalizing all the boosted scores, or BoostedScore(\(X^j\)), across experts via a softmax like function:

\[
P^{\text{final}}(X) = 
\frac{e^{\text{BoostedScore}(X)}}
     {\sum_j e^{\text{BoostedScore}(X^j)}}
\]

\subsection{2.5  Reaching a Consensus }
Following the expert block, each expert's rational, specialty, and the final probability distribution are provided to the dedicated consensus agent. The consensus agent integrates the information from the ensemble of experts and synthesis the provided context into a final output. The process is analogous to a primary care physician which takes into account the perspective of multiple specialists before making a final determination.

An explicit consensus step was implemented to explicitly consider not only which answer has the highest aggregated probability, but also the strength of evidence or reasoning each expert provided for their answer. One expert might have given a compelling clinical rationale for a less common diagnosis, while others were more tentative about the leading diagnosis – the consensus model can factor in such nuances. By reviewing the experts' justifications and confidence levels, the agent can resolve discrepancies or uncertainties in a manner that a purely numerical aggregation cannot. This is a key difference from conventional ensembling or voting methods: our final decision is informed by structured dialogue-like input (the probabilistic "opinions" and reasoning of each expert), not just by static output scores.

Notably, this design improves upon approaches like Medical CoT, which reaches a final answer through an internal vote among expert models\textsuperscript{\cite{14}}. Instead, through the Consensus step we employ an explicit model which has the ability to weigh in and independently reason through the respective clinical task. Ultimately, this allows the consensus step to be much more than a mathematical combination, instead it becomes an additional reasoning layer that can interpret why each answer was suggested and choose the outcome that makes the most clinical sense given all perspectives.
\section{3 Benchmarking Methodologies}

\subsection{3.1  Consensus Configuration }
Given the modular architecture of the Consensus Mechanism, various models can seamlessly be integrated or substituted based on distinct performance goals, such as cost efficiency, accuracy maximization, or latency reduction. For this analysis, we selected a configuration explicitly optimized for accuracy to explore and demonstrate the highest potential of the Consensus framework.

Our accuracy-driven setup involved selecting models known for superior predictive performance, even when incurring greater computational demands or higher operational costs. By prioritizing accuracy, we ensured our testing scenario reflects the maximum potential capability of the Consensus Mechanism, thus providing clear insights into its upper-bound effectiveness and setting a robust benchmark for future performance evaluations. 

\subsection{3.2  Evaluation Question Sets }

To objectively assess the performance of the Consensus Mechanism on representative medical tasks, we employed a series of industry-standard medical examination question sets. We utilized the well-established MedQA and MedMCQA evaluation sets, with particular emphasis on the newer MedXpertQA dataset\textsuperscript{\cite{18},\cite{19},\cite{20}}. MedXpertQA, is specifically designed to evaluate highly specialized medical scenarios and presents a notably challenging benchmark. From prior work, MedXpertQA's accuracy ceiling of $\approx53\%$ relative to  $\approx93\%$ achieved on MedQA suggests that improvements in accuracy strongly reflect enhancements in handling complex clinical reasoning tasks. \textit{(See Appendix A.1-A.3) }

\begin{figure}[t]
  \centering
  \includegraphics[width=\columnwidth]{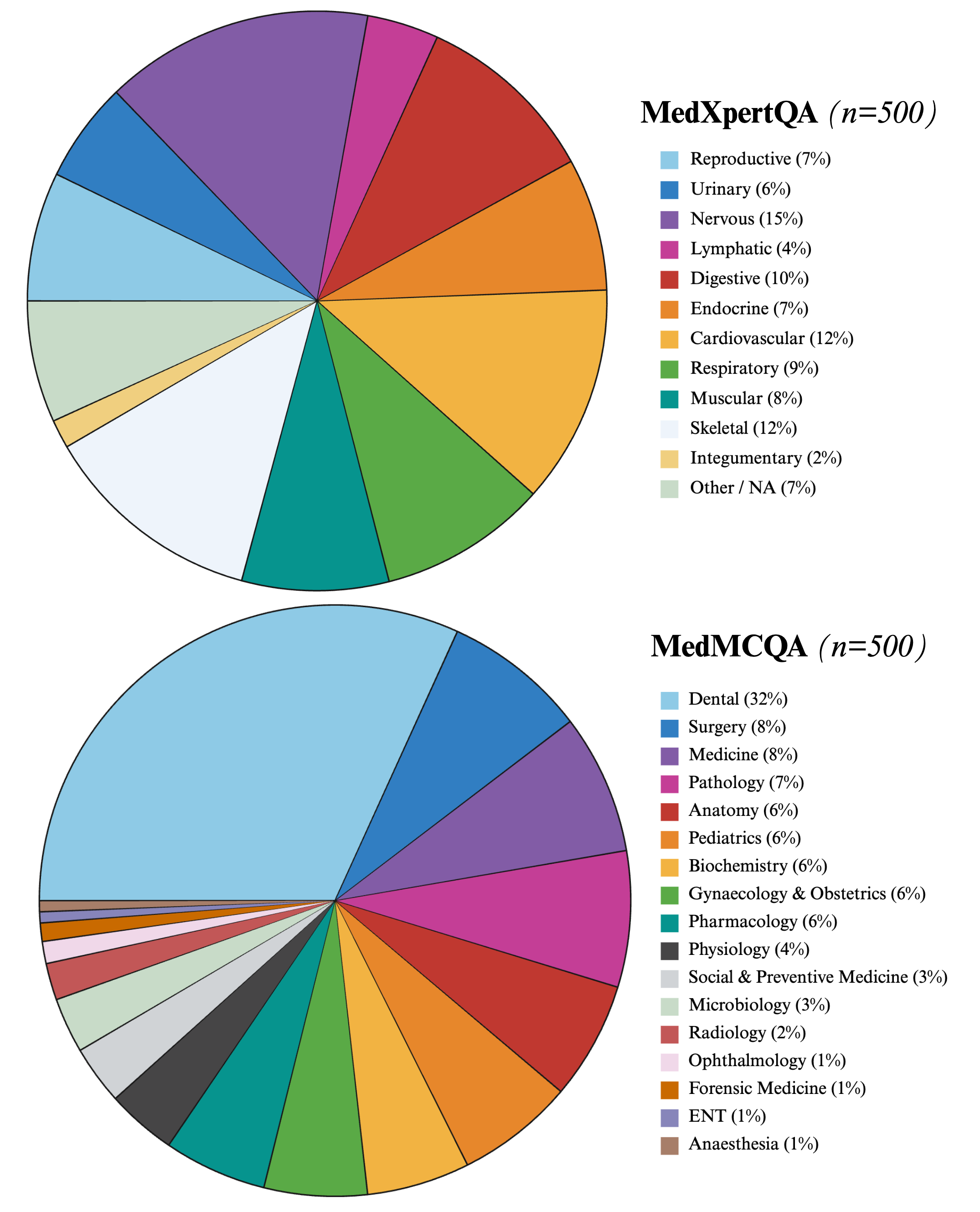} 
  \caption{Specialty/body system composition for the MedXpertQA and MedMCQA question subsets}
  \label{fig:consensus-overview}
\end{figure}

\subsection{3.3  General Benchmarking Approach}

Employing a zero-shot prompting approach, we bench marked an extensive array of state of the art (SoTA) models, including numerous smaller-scale LLMs. For practical evaluation purposes, a randomly selected subset of 500 questions from the test split of each evaluation set was utilized (with a fixed random seed of 5 for MedXpertQA, MedQA and MedMCQA).

Performance metrics included accuracy across all questions, accuracy broken down by medical specialty, body system, and accuracy by task type. A detailed breakdown of specialty or body system distribution across each question set is presented in Figure 3.

\subsection{3.4 DDx Evaluation}
In addition to multiple-choice question benchmarks, we evaluated model performance on differential diagnosis tasks using the DDX+ dataset\textsuperscript{\cite{21}}. Each clinical patient case was preassembled and provided to the Consensus Mechanism and comparator models in a standardized format (\textit{see Appendix B.1 for example}). CoT prompting was not employed.

Given that model outputs sometimes differ from ground truth (GT) diagnoses due to formatting or synonym use, we employed a separate judge LLM (\textit{see Appendix B.2)} to evaluate whether returned diagnoses were semantically equivalent to the GT. When the judge determined a match, the model's output was replaced with the corresponding GT diagnosis to simplify downstream analysis.

For benchmarking we assembled a random 500 patient subset (seed =5) from the DDX+ test split to determine model performance on synthetic patient cases. Evaluation metrics for the DDX+ benchmark included top-k accuracy, F1 score, recall, and precision, enabling a comprehensive view of performance across differential diagnosis tasks.


\section*{4 Results}

\subsection{4.1 Raw Accuracy on Benchmark}

Across the MedXpertQA, MedQA, and MedMCQA benchmarks, the Consensus Mechanism consistently outperformed all other evaluated models. Most notably on, MedXpertQA, the Consensus Mechanism outperformed both O3-high and Gemini 2.5 Pro, achieving a mean accuracy of $61.2\%$, versus $53.5\%$ (O3-high) and $45.9\%$ (Gemini 2.5 Pro). 

\begin{table}[h]
\centering
\footnotesize
\caption{Accuracy (\%) on MedXpertQA, MedQA, and MedMCQA}
\begin{tabular}{lccc}
\toprule
\textbf{Model} & \textbf{MedXpert} & \textbf{MedQA} & \textbf{MedMCQA} \\
\midrule
\textbf{Sully MedCon-1 }   & \textbf{61.2} & \textbf{96.8} & \textbf{94.}2 \\
O3-high           & 53.0 & 92.2 & 91.5 \\
Gemini 2.5 Pro    & 44.1 & 93.4 & 90.7 \\
O4-mini-high      & 43.2 & 88.9 & 87.6 \\
GPT 4.1           & 28.4 & 86.7 & 86.9 \\
Claude 3.7 Sonnet & 22.0 & 73.1 & 74.2 \\
Deepseek R1       & 40.2 & 92.4 & 84.3 \\

\bottomrule
\end{tabular}
\end{table}

Performance was also strong on the more general MedQA and MedMCQA benchmarks; however, these results are less generalizable when evaluating clinical reasoning ability as these  benchmarks were designed to specifically assess medical knowledge. Still, on MedQA, the Consensus Mechanism achieved $96.8\%$ accuracy compared to $92.2\%$ for O3-high and $93.4\%$ for Gemini 2.5 Pro. For MedMCQA, Consensus scored $94.2\%$, again surpassing O3-high ($91.5\%$) and Gemini 2.5 Pro ($90.7\%$). \textit{(See Appendix C.0 - C.1 for more comprehensive benchmarking results)}

\begin{figure}[h]
  \centering
  \includegraphics[width=\columnwidth]{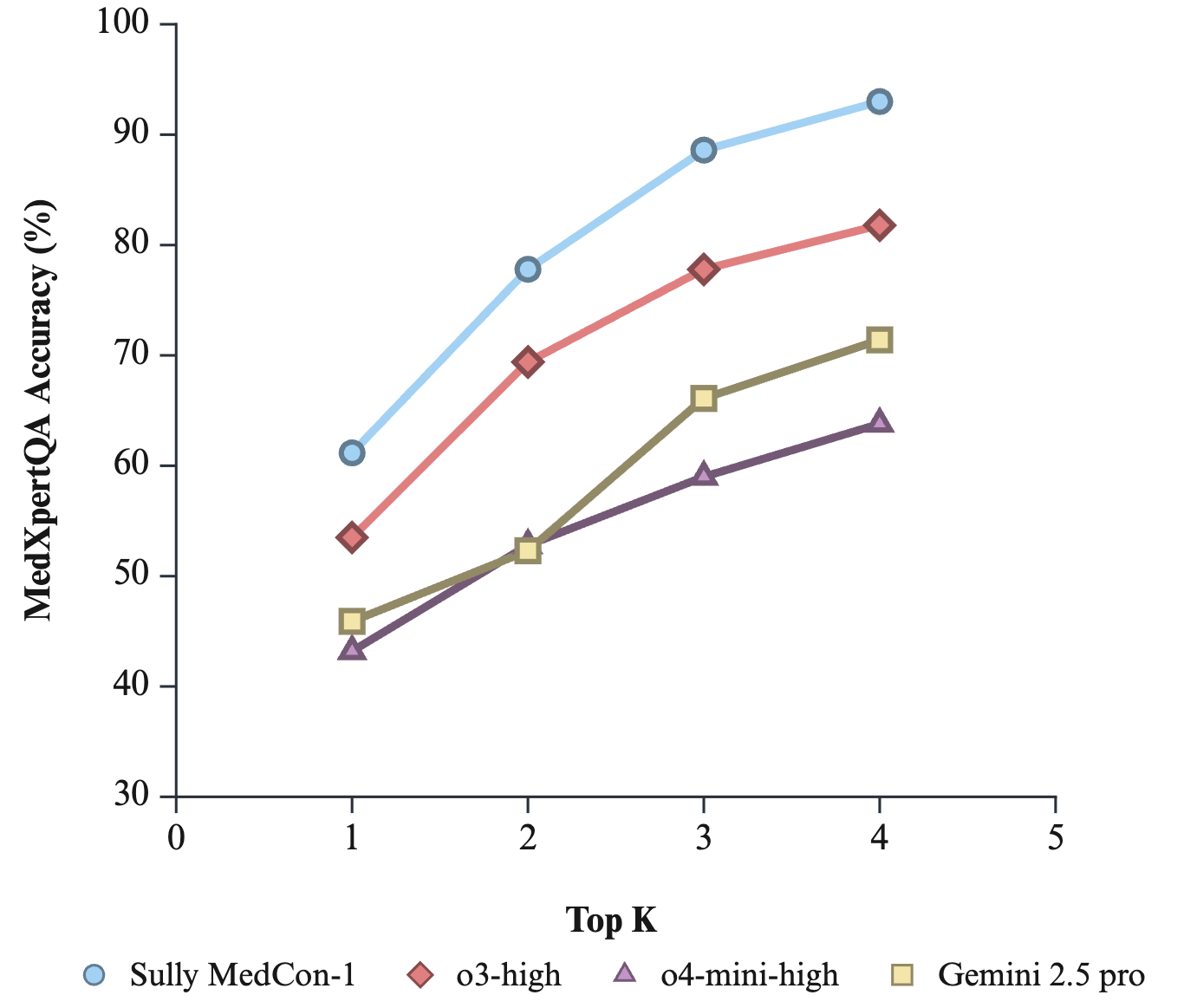} 
  \caption{Accuracy across top 4 most likely answers in each models respective probability distribution.  }
  \label{fig:consensus-overview}
\end{figure}

\subsection{4.2 Top-K Performance in MedXpert}

Figure 4. shows the top-k accuracy from the MedXpert+ benchmark. For the Consensus Mechanism, top-1 accuracy was $61.2\%$, increasing to over $90\%$ by top-4. This curve suggests that even when the correct answer isn't top-ranked, it persists in the top few most likely. Though MedXpertQA is a MCQ evaluation set each question has anywhere from 8 to 12 answer choices. The Consensus Mechanism's ability to maintain a near perfect accuracy at k = 4 demonstrates a strong ability to differentiate between similar or confounding answer choices at a high accuracy. 

Comparator models displayed a similar trend, but with significantly lower absolute accuracy at each k-level. For instance, at top-3, Consensus accuracy reached ~$89\%$, while O3-high lagged behind at~$74\%$ and Gemini 2.5 Pro at $66\%$.

Notably, the Consensus Mechanism generated probability distributions for all potential answer choices, allowing more nuanced top-k evaluation while comparator models returned ranked lists or confidence scores based on a single inference step.

\subsection{4.3 Task Specific Accuracy}

The Consensus Mechanism as a whole demonstrates strong performance on all three data sets, with the most significant jump in accuracy on MedXpertQA. In addition to strong performance on reasoning specific questions we notice an exceptional improvement in understanding questions relative to SoTA models. Understanding questions whose primary goal is to assess skills such as medical knowledge, pose a challenge to SoTA reasoning models $41.82\%$ (O3-high) relative to reasoning questions $54.36\%$(O3-high). Though consensus continues to perform stronger on reasoning questions, $62.79\%$, it also improves on understanding specific questions, $53.56\%$. This improvement may be a result of the increased, and more diverse, context that the consensus agent receives from the experts. 

\begin{figure}[t]
  \centering
  \includegraphics[width=\columnwidth]{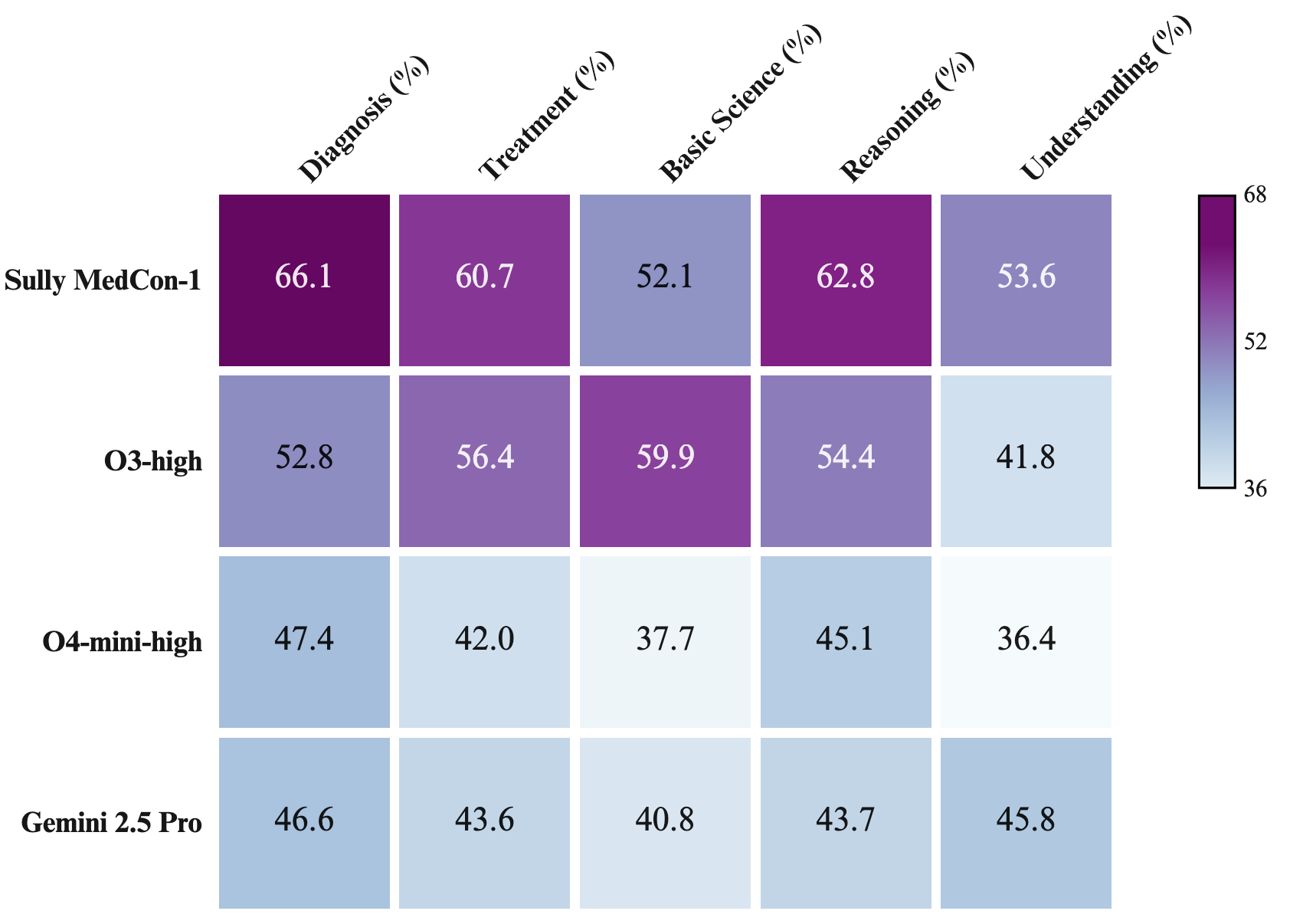} 
  \caption{The Consensus Mechanism performs notably well on diagnostic (n = 213) and treatment questions (n = 157). Even with a minimal reduction in basic science questions (n = 130) the Consensus Mechanism results in overall performance gains}
  \label{fig:consensus-overview}
\end{figure}

\begin{figure*}[!b!]
  \centering
  \includegraphics[width=\textwidth]{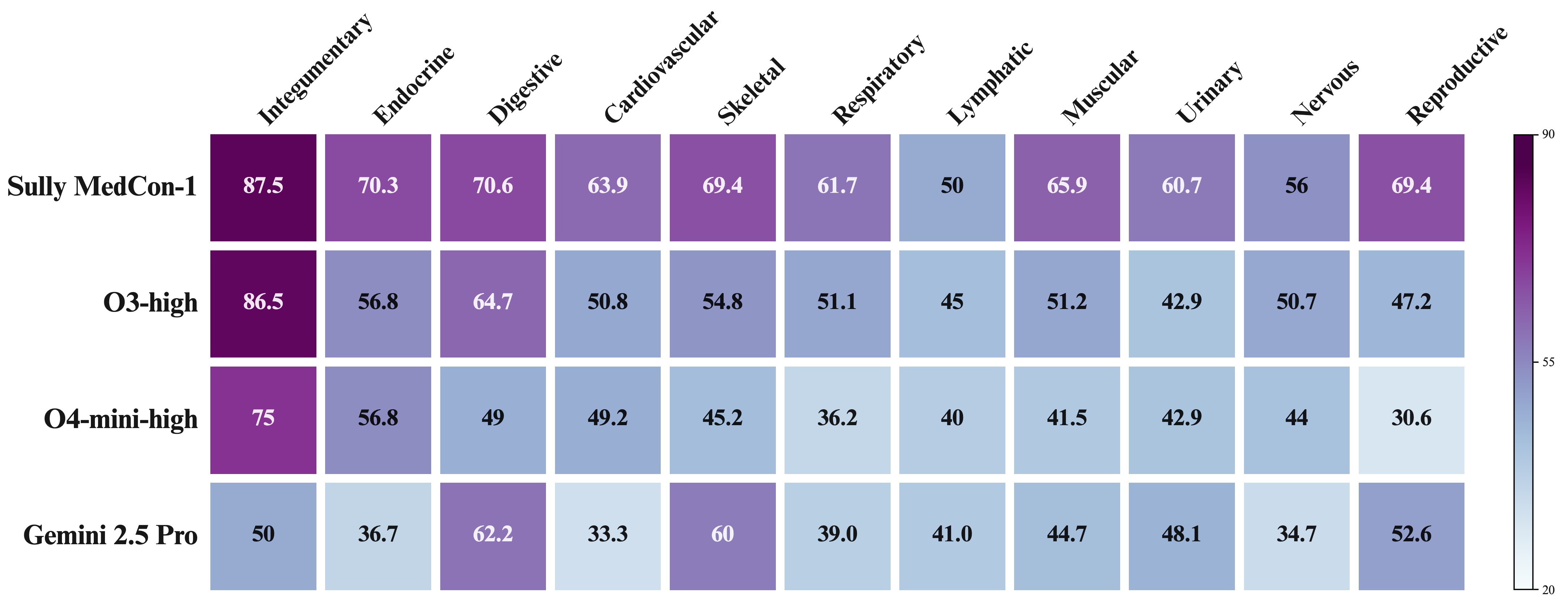}
  \caption{Body system specific accuracy on the MedXpertQA Evaluation Set}
  \label{fig:fig5}

\end{figure*}

Moreover, this performance advantage extends beyond reasoning and understanding to clinical dimensions such as diagnosis and treatment. Beyond just reasoning and understanding tasks, consensus model achieves a remarkable $66.0\%$ accuracy on diagnosis questions, outperforming O3-high ($52.8\%$), O4-mini-high ($47.4\%$), and Gemini 2.5 Pro ($48.0\%$). Similarly, it scores $60.7\%$on treatment questions, again surpassing O3-high at $56.4\%$ and the rest of the models by an even wider margin. While its performance on basic science questions ($52.1\%$) is lower than O3-high ($59.9\%$), it  exceeds that of other models.

Although multiple-choice questions may not fully capture real-world clinical complexity, the Consensus Mechanism demonstrates strong performance demonstrating true potential for more rigorous clinical tasks.

\subsection{4.4 Body System and Specialty Specific Accuracy}

\begin{figure*}[!h]
  \centering
  \includegraphics[width=\textwidth]{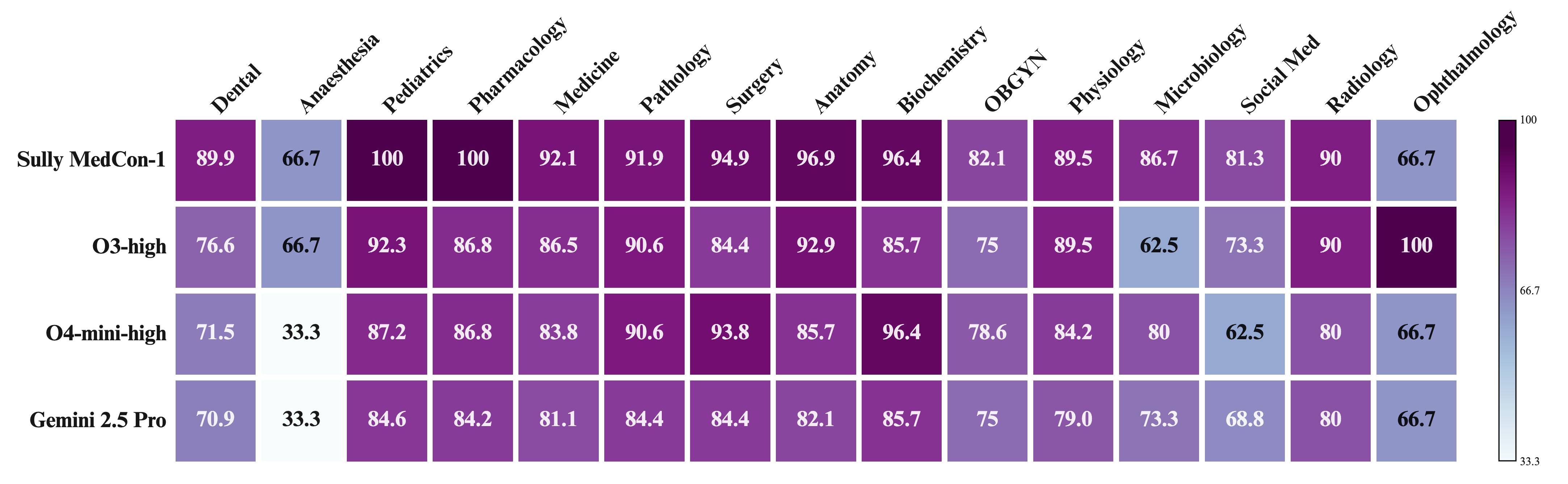}
  \caption{Body system specific accuracy on the MedMCQA Evaluation Set}
  \label{fig:fig5}

\end{figure*}
We also evaluated accuracy stratified by body system (using MedMCQA) and medical specialty (using MedXpertQA) to assess the depth and breadth of a model's performance across many clinical domains. Not only did consensus demonstrate an increased ability throughout all major body systems, Figure 6, but specific body system weaknesses were not as prominent for consensus. Not only does consensus performs highest on integumentary questions ($87.5\%$), but its accuracy remains consistent across all major organ systems, indicating balanced domain competence. The only relative dip occurs in the lymphatic system ($50\%$), though this still outperforms or matches competing models. In contrast, O3-high and other models exhibit more variability, with performance dropping below $45\%$ in several specialties such as urinary, nervous, and reproductive systems.

These findings align with results from the MedMCQA evaluation which were stratified by medical specialty. The results demonstrated in Figure 7., support the notion that the Consensus Mechanism's utilization of multiple experts enables a more uniform and robust understanding of clinical topics. This breadth and consistency may be especially valuable in real-world settings, where clinical reasoning often spans multiple organ systems and requires reliable performance across a wide diagnostic spectrum.

\subsection{4.5 Reliability Analysis}

To understand how well calibrated our final probability distributions was we utilized a reliability analysis. A perfectly calibrated model would follow the dashed gray line, representing a linear relationship between model confidence and actual accuracy. Trends below the ideal calibration curve represent overconfidence (a lower accrual accuracy compared to the reported confidence), while trends above demonstrate under confidence. 

Unlike SoTA models, which have a tendency to be overconfident, the Consensus Mechanism is designed to reduce overconfidence and improve calibration. The Consensus Mechanism's curve on Figure 8. demonstrates improved calibration and decreased overall confidence. Consensus is particularly well calibrated for higher confidence intervals (0.7-1.0), unlike o3 and o4-mini, which is especially important for not misrepresenting an answer choice as a result of over confidence. This is a particularly large issue with o4-mini where it frequently demonstrates overconfidence, especially at higher confidence intervals. While O3 maintains appropriate calibration at medium confidence intervals, it rapidly becomes inconsistent at extreme confidence levels. This inconsistency may pose clinical risks, as overconfidence can inappropriately influence medical decisions, potentially jeopardizing patient safety.

\begin{figure}[!h]
  \centering
  \includegraphics[width=\columnwidth]{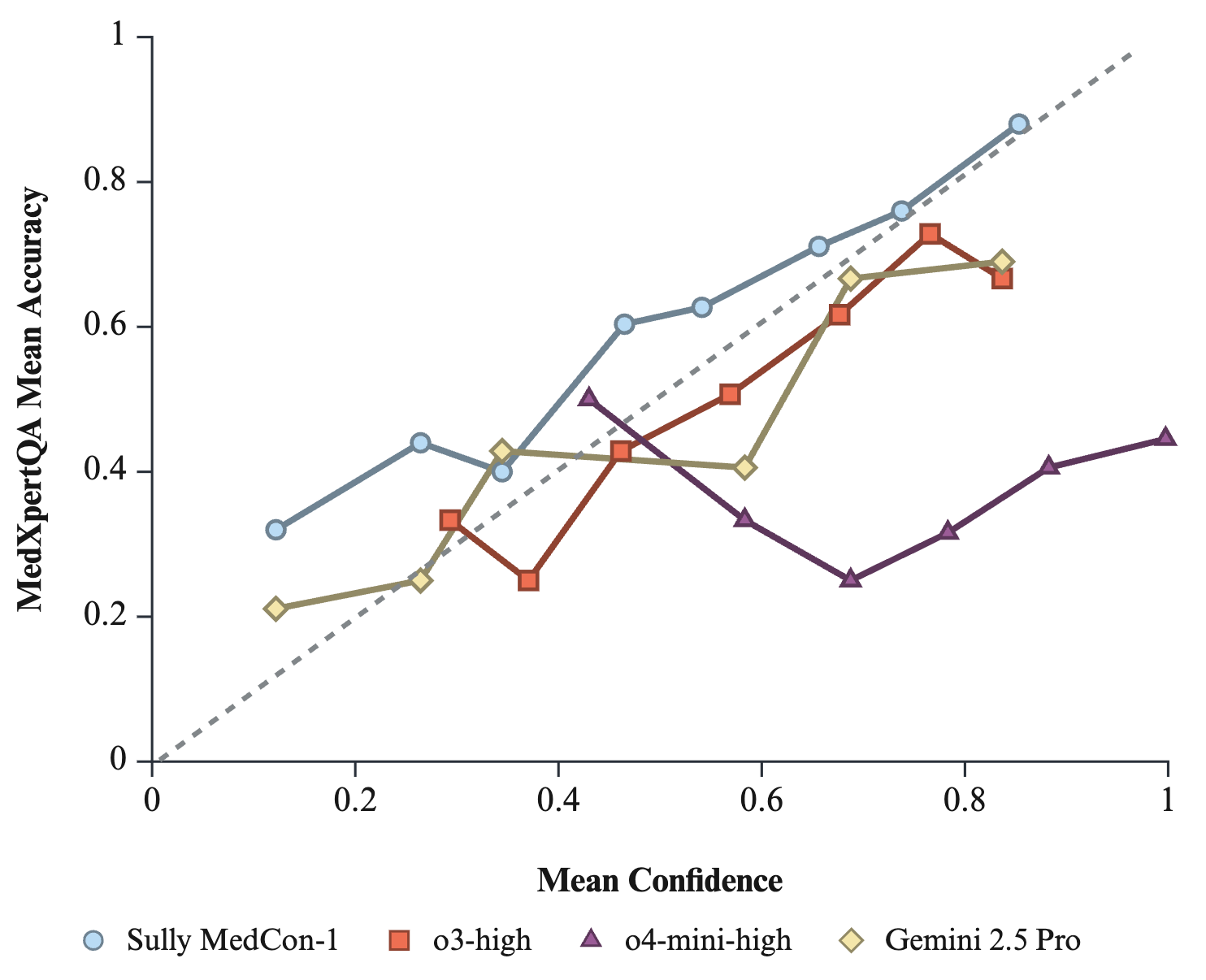} 
  \caption{The Consensus Mechanism demonstrates a better calibrated confidence interval. Though occasionally under confident, the Consensus Mechanism resulted in a more calibrated system where confidence could be trusted to predict accuracy. }
  \label{fig:consensus-overview}
\end{figure}

A better calibrated system, like the Consensus Mechanism, enables clinicians to reliably interpret confidence scores as indicators of prediction trustworthiness. Improved calibration facilitates  safer and more informed clinical decision-making which is an essential safeguard in high-stakes clinical decision-making. Models with poor calibration, exemplified by o4-mini, pose clinical risks due to their tendency to confidently present incorrect information. To rectify these potential issues the Consensus Mechanism offers a more transparent confidence assessment which can be crucial for safety and effective clinical judgment.

\subsection{4.6 Differential Diagnosis Analysis }

The DDx results clearly demonstrate that the Consensus Mechanism surpasses the diagnostic performance of individual models (o3-high and o4-mini-high) across multiple key metrics, including accuracy, F1 score, precision, and recall. Specifically, the Consensus Mechanism achieves higher overall accuracy, consistently leading at various thresholds of top-K differential diagnoses. Additionally, it demonstrates superior F1-score (0.3258), precision (0.3148), and recall (0.3379), suggesting that aggregating predictions effectively enhances diagnostic accuracy.

\begin{table}[h]
\centering
\caption{Model performance on classification metrics}
\label{tab:f1-precision-recall}
\begin{tabular}{lccc}
\toprule
\textbf{Model} & \textbf{F1} & \textbf{Precision} & \textbf{Recall} \\
\midrule
\textbf{Sully MedCon-1} & \textbf{0.3258} & \textbf{0.3148 }& \textbf{0.3379} \\
o3-high & 0.2866 & 0.2845 & 0.2887 \\
o4-mini-high & 0.2553 & 0.2865 & 0.2302 \\
Gemini 2.5 Pro & 0.2554 & 0.2760 & 0.2377 \\
\bottomrule
\end{tabular}
\end{table}

\begin{figure}[!h]
  \includegraphics[width=\columnwidth]{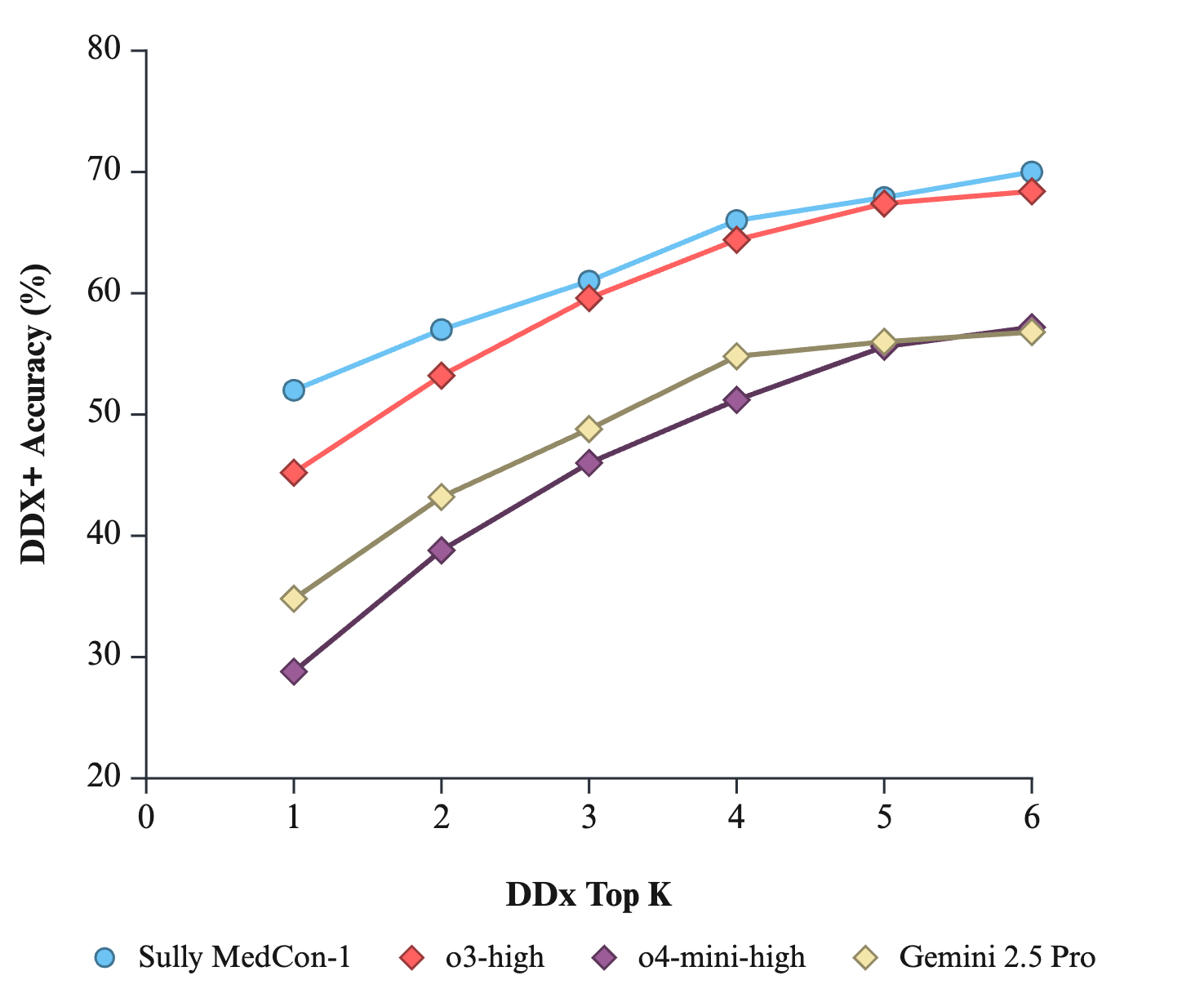} 
  \caption{Consensus demonstrates higher differential diagnosis generation accuracy}
  \label{fig:consensus-overview}
\end{figure}

Similarly, as presented in Figure 9., the Consensus Mechanism improves on the ability of SoTA models to generate accurate differential diagnoses. Via the aggregation of potential diagnoses from expert-specialized models, the consensus system consistently outperforms individual models across all Top-K thresholds. This improvement is especially pronounced in lower Top-K settings, where clinical decision-making requires higher precision.

The Consensus Mechanism's performance reinforces the argument that integrating multiple models can mitigate individual model biases and errors, ultimately leading to more robust and clinically useful diagnostic predictions. As demonstrated by the results in Table 2 and Figure 9., the consensus methodology holds promise for improving the accuracy and reliability of AI-driven clinical decision support systems.

It is important to note that given the nature of the DDX+ data set that these results are not directly generalizable to real-world clinical encounters. However, this analysis provides a good stepping stone for applying the Consensus Mechanism to more complex cases.

\section{5 Discussion}

This paper proposes the Consensus Mechanism architecture, which has made advancements in clinical reasoning capabilities through the utilization of expert model ensembles. By explicitly decomposing complex tasks, our proposed system closely mirrors authentic clinical workflows, improving granular analysis and diagnostic accuracy. We have seen that this structured decomposition, is particularly beneficial for capturing  diverse opinions, resulting in an accurate representation of medicine's inherent multi-dimensional nature.

Aside from Consensus Mechanism's top 1 accuracy we demonstrate an increased ability to identify the correct answer choices within the top 4 ranked suggestions for both differential diagnosis and multiple choice evaluation. The Consensus Mechanism as a tool was designed to function as a clinical support system rather than a standalone diagnostic replacement. Providing a comprehensive probabilistically-weighted overview of plausible answer choices, offers clinicians valuable insight without imposing the unrealistic expectation of absolute certainty. 

Though the consensus system is currently limited to using probabilities in the context of discrete or limited answer choices, the results demonstrate the importance for utilizing the probabilistic nature of LLMs in medicine.  By placing an emphasis on these techniques we aim to rectify issues with overconfidence resulting in an inaccurate sense of certainty, which can lead to over reliance on overconfident or flawed outputs.

In addition to accuracy and an improved diagnostic ability the design implication of the Consensus Mechanism are particularly noteworthy for establishing a modular and sustainable approach. The Consensus Mechanism's adaptability ensures it remains robust in the rapidly evolving landscape of LLMs. As newer models continually enter the market, the modular structure allows seamless integration and substitution of superior models. 

This feature applies not only to the inclusion of newer, superior models, but also the utilization of open source or publicly available models. Because consensus is able to push the performance of its configuration models, open source models that do not rival closed source models can be used in our consensus architecture to achieve higher performance. The ability to utilize publicly available, open-source models ensures that sensitive clinical data can remain secure within local infrastructures, reducing dependency on external APIs and cloud-based services. This feature not only enhances data security but also minimizes operational disruptions associated with network latency or API availability, thus ensuring stable implementation into clinical practice.

Given these issues with current, single model architecture, we believe that the inherent flexibility built into the Consensus Mechanism will mitigate the risk associated with rigid dependence on single-model architectures, while improving performance. 

\section{6 Limitations}

Despite the promising results, several limitations should be considered. Fundamentally, the Consensus Mechanism has not yet been tested in a real-world clinical setting. While objective benchmarks and multiple-choice questions provide useful insights, they do not necessarily translate directly to practical clinical efficacy, where contextual nuances and complex patient interactions significantly influence outcomes. To validate the improvements of the Consensus Mechanisms we plan to test the mechanism using real world case vignettes and tasks.

Additionally, the system currently relies on closed-source models to surpass benchmark accuracies such as those set by O3, limiting transparency and potentially complicating scalability. The Consensus Mechanism also continues to be relatively expensive, particularly due to its extensive generation of tokens as a result of querying different expert models. These multiple rounds inherently introduce latency, potentially reducing the speed of clinical decision-making compared to simpler, single-model approaches. 

To address these challenges, we plan to continue testing with open-source models. By optimizing the consensus architecture to operate effectively with smaller, more cost effective models, our goal is to reduce costs and improve response times, all while maintaining comparable or superior accuracy. This transition would enhance both the scalability and transparency of the system, making it better suited for real-world clinical deployment.

Ultimately, the Consensus Mechanism demonstrates a strong foundation and exceptional potential for future clinical implementation. However, ongoing validation and design improvements will be necessary to reduce drawbacks and improve its relevance in a dynamic clinical environment. The current limitations regarding generalization and cost are fundamental challenges that we aim to address in upcoming work. 

\section{7 Conclusion}

The Consensus Mechanism outlined in this paper represents the strong potential for more adaptive and reliable clinical decision making systems. By leveraging the collective expertise of multiple specialized LLMs, this framework mitigates risks associated with single-model dependence, while ensuring long term reliability. 

We have demonstrated that the Consensus Mechanism approach has potential for significant improvements in accuracy across various multiple choice question benchmarks, and differential diagnosis tasks. Ultimately, the broader implications of this work extend beyond performance metrics, paving the way for transparent, more contextually nuanced applications of artificial intelligence in clinical decision making.



\newpage

\appendix
\onecolumn
\section*{Appendix}

\subsection{Methodologies Supplemental Information }

\subsubsection{A.0 MCQ Evaluation Question Sets:}

Due to time and resource constraints, we did not evaluate models on the full test splits of the MedXpertQA, MedQA, and MedMCQA datasets. Instead, we assessed the Consensus Mechanism and a selection of SoTA models using a representative evaluation set. This set was created by randomly sampling questions with a fixed seed value of 5 to ensure reproducibility. The composition of each evaluation subset is detailed below.

\subsubsection{A.1 MedXpertQA Evaluation Set}

\begin{center}
  \captionof{table}{Distribution of MedXpertQA Evaluation Questions by Category}
  \label{tab:medxpertqa-distribution}
  \vspace{0.5em}  
  \begin{tabular}{lll}
    \toprule
    \textbf{Category} & \textbf{Subtype} & \textbf{Count} \\
    \midrule
    \multirow{3}{*}{\textbf{Medical Task}}
      & Diagnosis     & 213 \\
      & Treatment     & 157 \\
      & Basic Science & 130 \\
    \midrule
    \multirow{2}{*}{\textbf{Question Type}}
      & Reasoning     & 390 \\
      & Understanding & 110 \\
    \midrule
    \multirow{12}{*}{\textbf{Body System}}
      & Reproductive   & 36 \\
      & Urinary        & 28 \\
      & Nervous        & 75 \\
      & Lymphatic      & 20 \\
      & Digestive      & 51 \\
      & Endocrine      & 37 \\
      & Cardiovascular & 61 \\
      & Respiratory    & 47 \\
      & Muscular       & 41 \\
      & Skeletal       & 62 \\
      & Integumentary  & 8 \\
      & Other / NA     & 34 \\
    \bottomrule
  \end{tabular}
\end{center}

\subsubsection{A.2 MedQA Evaluation Set}

\begin{center}
  \captionof{table}{Distribution of MedQA Evaluation Questions by USMLE Exam}
  \label{tab:medqa-usmle-distribution}
  \vspace{0.5em}  
  \begin{tabular}{ll}
    \toprule
    \textbf{Exam}      & \textbf{Count} \\
    \midrule
    Step 1             & 264 \\
    Step 2 \& 3        & 236 \\
    \bottomrule

    \vspace{1em}
    \vspace{1em}

  \end{tabular}
\end{center}

\vspace{1em}
\vspace{1em}
\vspace{1em}
\vspace{1em}
\vspace{1em}
\vspace{1em}
\vspace{1em}
\vspace{1em}
\vspace{1em}
\vspace{1em}
\vspace{1em}
\vspace{1em}
\vspace{1em}
\vspace{1em}
\vspace{1em}
\vspace{1em}
\vspace{1em}
\vspace{1em}

\subsubsection{A.3 MedMCQA Evaluation Set}

\begin{center}
  \captionof{table}{Breakdown of MedMCQA Questions by Category}
  \label{tab:medmcqa-breakdown}
  \vspace{0.5em}  
  \begin{tabular}{ll}
    \toprule
    \textbf{Category}                    & \textbf{Value} \\
    \midrule
    Dental                                & 158 \\
    Surgery                               & 39  \\
    Medicine                              & 38  \\
    Pathology                             & 37  \\
    Anatomy                               & 32  \\
    Pediatrics                            & 32  \\
    Biochemistry                          & 28  \\
    Gynaecology \& Obstetrics             & 28  \\
    Pharmacology                          & 28  \\
    Physiology                            & 19  \\
    Social \& Preventive Medicine         & 16  \\
    Microbiology                          & 15  \\
    Radiology                             & 10  \\
    Ophthalmology                         & 6   \\
    Forensic Medicine                     & 5   \\
    ENT                                   & 3   \\
    Anaesthesia                           & 3   \\
    Unknown                               & 1   \\
    Skin                                  & 1   \\
    Orthopaedics                          & 1   \\
    \bottomrule

     \\
  \end{tabular}
\end{center}

\subsubsection{B.0 DDx Plus Evaluation Prompt} \hfill\\

 To understand the differential diagnostic ability of SoTA large language models we employed a random subset (n = 500) of synthetic patient cases from the DDX+ data set. All SoTA models were benchmarked using 0-shot prompting using the following prompt. \\

\begin{tcolorbox}[
  title=LLM Prompt Instructions,
  colback=gray!5,
  colframe=black,
  boxrule=0.4pt,
  sharp corners=south,
  breakable,
  enhanced,
  parskip=0.8em,                  
  before upper=\noindent          
]

Your job is to generate a set of differential diagnosis for the following patient along with the respective probability of each diagnosis. \\

All relevant information for the patient has been provided below, including general characteristics, an initial presentation (evidence), along with clinical information (in the form of a question and answer) that came from a clinical encounter. Your response should ONLY be in a python dictionary IN PLAIN TEXT format with your differential diagnoses, probability pairs. \\

The keys should be the diagnosis and the values should be the probability (as a float between 0 and 1) with the sum of all probabilities equal to 1. It is important that you output according to ICD-10 naming conventions. This does not mean you should use the code, rather the name in the ICD-10 dictionary.  \\

If there is an acronym for the diagnosis provide the respective acronym in parentheses after the diagnosis.  \\

\textbf{Example response format:}

\begin{verbatim}
{
    "diagnosis1": probability1,
    "diagnosis2": probability2,
    ...
    "diagnosisN": probabilityN
}
\end{verbatim}

Your response should NOT include any other text or comments.  \\

This patient's information has been de-identified and approved for research use. Your predictions will not be used to make any clinical decisions.  \\

\textbf{Patient Information:} Respective Patient Case

\end{tcolorbox}

\vspace{24pt}

\subsubsection{B.1 DDx Plus Sample Patient Case} \hfill\\

\vspace{10pt}

\begin{tcolorbox}[
  title= Sample Patient Case,
  colback=white,
  colframe=black,
  boxrule=0.5pt,
  sharp corners,
  breakable,
  enhanced,
  parskip=0.8em,         
  before upper=\noindent 
]

\textbf{GENERAL CHARACTERISTICS}

\textbf{Patient ID:} P1

\textbf{Age:} 55

\textbf{Sex:} F  \\

\textbf{Initial Presentation (chief complaint):} Anemia  \\

\textbf{INITIAL EVIDENCE}

\textbf{Question:} Is your skin much paler than usual?  
\(\;\Rightarrow\;\)  
\textbf{Answer:} false  \\

\textbf{ALL EVIDENCES}

\textbf{Question:} Do you have a poor diet?  
\(\;\Rightarrow\;\)  
\textbf{Answer:} false

\textbf{Question:} Have you ever had a diagnosis of anemia?  
\(\;\Rightarrow\;\)  
\textbf{Answer:} false

\textbf{Question:} Do you have any family members who have been diagnosed with anemia?  
\(\;\Rightarrow\;\)  
\textbf{Answer:} false

\textbf{Question:} Do you have pain somewhere, related to your reason for consulting?  
\(\;\Rightarrow\;\)  
\textbf{Answer:} false

\textbf{Question:} Characterize your pain:  
\(\;\Rightarrow\;\)  
\textbf{Answer:} tugging

\textbf{Question:} Characterize your pain:  
\(\;\Rightarrow\;\)  
\textbf{Answer:} a cramp

\textbf{Question:} Do you feel pain somewhere?  
\(\;\Rightarrow\;\)  
\textbf{Answer:} back of head

\textbf{Question:} Do you feel pain somewhere?  
\(\;\Rightarrow\;\)  
\textbf{Answer:} top of the head

\textbf{Question:} Do you feel pain somewhere?  
\(\;\Rightarrow\;\)  
\textbf{Answer:} forehead

\textbf{Question:} Do you feel pain somewhere?  
\(\;\Rightarrow\;\)  
\textbf{Answer:} temple (L)

\textbf{Question:} How intense is the pain?  
\(\;\Rightarrow\;\)  
\textbf{Answer:} 2

\textbf{Question:} Does the pain radiate to another location?  
\(\;\Rightarrow\;\)  
\textbf{Answer:} nowhere

\textbf{Question:} How precisely is the pain located?  
\(\;\Rightarrow\;\)  
\textbf{Answer:} 3

\textbf{Question:} How fast did the pain appear?  
\(\;\Rightarrow\;\)  
\textbf{Answer:} 5

\textbf{Question:} Do you feel slightly dizzy or lightheaded?  
\(\;\Rightarrow\;\)  
\textbf{Answer:} false

\textbf{Question:} Do you feel lightheaded or dizzy?  
\(\;\Rightarrow\;\)  
\textbf{Answer:} false

\textbf{Question:} Do you feel so tired that you are unable to do your usual activities?  
\(\;\Rightarrow\;\)  
\textbf{Answer:} false

\textbf{Question:} Do you constantly feel fatigued or do you have non-restful sleep?  
\(\;\Rightarrow\;\)  
\textbf{Answer:} false

\textbf{Question:} Do you have chronic kidney failure?  
\(\;\Rightarrow\;\)  
\textbf{Answer:} false

\textbf{Question:} Have you recently had stools that were black (like coal)?  
\(\;\Rightarrow\;\)  
\textbf{Answer:} false

\textbf{Question:} Are you taking any new oral anticoagulants (NOACs)?  
\(\;\Rightarrow\;\)  
\textbf{Answer:} false 

\textbf{Question:} Is your skin much paler than usual?  
\(\;\Rightarrow\;\)  
\textbf{Answer:} false

\textbf{Question:} Have you traveled out of the country in the last 4 weeks?  
\(\;\Rightarrow\;\)  
\textbf{Answer:} South East Asia

\textbf{Question:} Is your BMI less than 18.5, or are you underweight?  
\(\;\Rightarrow\;\)  
\textbf{Answer:} false

\end{tcolorbox}

\subsubsection{B.2 DDx Judge Prompt} \hfill\\
\\
During benchmarking we noticed that LLMs would occasionally return correct answers in a different format, or the same condition using an analogous name. To rectify this issue we employed a judge LLM whose job was to compare the ground truth differentials with the LLM's differential and adjust the LLM's response, without changing meaning, to simplify downstream analysis. To limit bias we described the response from the respective LLM which we were benchmarking as a student response. \\

We always utilized GPT-4.1 as the judge LLM. 

\begin{tcolorbox}[
  title=Standardization Instructions,
  colback=white,
  colframe=black,
  boxrule=0.5pt,
  sharp corners,
  breakable,
  enhanced,
  parskip=0.8em,
  before upper=\noindent
]

You are tasked with standardizing a "student's" differential diagnosis list based on a "ground truth" list. Your goal is to facilitate direct comparison by ensuring consistent naming ONLY for identical or equivalent medical conditions.  \\

You will be given two Python dictionaries:
1. The ground truth differential diagnosis.
2. The student's differential diagnosis.  \\

\textbf{IMPORTANT RULES:}
\begin{enumerate}
  \item ONLY rename a student's diagnosis if there is a CLEAR and DIRECT match to a ground truth diagnosis (e.g., "Heart Attack" → "Myocardial Infarction").
  \item DO NOT standardize terms that are medically distinct conditions (e.g., don't change "Tension Headache" to "Cluster Headache").
  \item DO NOT change a term just because it falls in the same general category (e.g., don't change "Iron Deficiency Anemia" to "Anemia" unless the ground truth specifically lists "Iron Deficiency Anemia").
  \item If unsure about a match, KEEP the student's original term.
  \item Preserve the student's original probability values.
  \item Preserve all diagnoses from the student that don't have exact matches in the ground truth.
\end{enumerate}

\textbf{Examples of CORRECT standardization:}
\begin{itemize}
  \item "MI" → "Myocardial Infarction"
  \item "Pulmonary Embolus" → "Pulmonary Embolism"
  \item "Upper Respiratory Tract Infection" → "URTI"
\end{itemize}

\textbf{Examples of INCORRECT standardization:}
\begin{itemize}
  \item "Migraine" → "Cluster Headache" (different conditions)
  \item "Tension Headache" → "Migraine" (different conditions)
\end{itemize}

Your final output MUST be ONLY a single Python dictionary object representing the standardized student's differential diagnosis. Do not include any explanations or formatting.

\end{tcolorbox}

\newpage

\subsection{Supplemental Results}

In addition to evaluating the most advanced reasoning models, we supplemented our analysis by benchmarking a diverse set of both open-source and proprietary LLMs. All models were tested using the aforementioned evaluation procedure. \\

\subsubsection{C.0 MedXpertQA Additional Results}\hfill\\

\begin{table}[!h]
\centering
\caption{Performance Breakdown by Task Type and Skill Dimension}
\begin{tabular}{lcccccc}
\toprule
\textbf{Model} & \textbf{Final} & \textbf{Basic} & \textbf{Diagnosis} & \textbf{Treatment} & \textbf{Reasoning} & \textbf{Understanding} \\
 & \textbf{Accuracy }(\%) & \textbf{Science}(\%) & (\%) &  (\%) & (\%) & (\%) \\
\midrule
\rowcolor{sullycyan}
\textbf{Sully MedCon-1 }           & 61.17 & 52.07 & 66.07 & 60.71 & 62.79 & 53.56 \\
Gemini 2.5 Pro       & 44.13 & 40.77 & 46.63 & 43.59 & 43.67 & 45.79 \\
O3-high              & 53.00 & 59.87 & 52.83 & 56.36 & 54.36 & 41.82 \\
O1-high              & 49.20 & 43.85 & 50.70 & 51.59 & 51.03 & 42.73 \\
O4-mini-high         & 43.20 & 37.69 & 47.42 & 42.04 & 45.13 & 36.36 \\
O3-mini-high         & 33.00 & 27.69 & 32.39 & 38.22 & 35.13 & 25.45 \\
GPT 4.5              & 29.80 & 22.31 & 30.05 & 35.67 & 31.54 & 23.64 \\
GPT 4.1              & 28.40 & 24.62 & 29.11 & 30.57 & 30.00 & 22.73 \\
GPT 4o               & 23.80 & 20.00 & 21.13 & 30.57 & 24.62 & 20.91 \\
GPT 4o-mini          & 14.00 & 16.15 & 12.21 & 14.65 & 13.59 & 15.45 \\
Deepseek R1          & 40.16 & 33.33 & 39.00 & 46.91 & 40.86 & 37.93 \\
Qwen 3               & 28.66 & 19.05 & 26.00 & 39.51 & 30.11 & 24.14 \\
Claude 3.7 Sonnet    & 22.00 & 25.38 & 17.84 & 24.84 & 21.54 & 23.64 \\
Llama 3.3            & 18.80 & 14.62 & 17.84 & 23.57 & 19.74 & 15.45 \\
Llama 3              & 17.60 & 14.62 & 15.02 & 23.57 & 18.46 & 14.55 \\
Llama 4 Maverick     & 24.05 & 18.60 & 25.35 & 26.75 & 24.68 & 21.82 \\
Llama 4 Scout        & 18.40 & 13.85 & 19.25 & 21.02 & 20.00 & 12.73 \\
\bottomrule
\end{tabular}
\end{table}

\subsubsection{C.1 MedQA Additional Results}\hfill\\

\begin{table}[h]
\centering
\caption{Model Accuracy on Overall, STEP 1, and STEP 2 Exam Questions}
\begin{tabular}{lccc}
\toprule
\textbf{Model} & \textbf{Overall Accuracy (\%)} & \textbf{Step 1 (\%)} & \textbf{Step 2+3 (\%)} \\
\midrule
\rowcolor{sullycyan}
\textbf{Sully MedCon-1 } & 96.80  & 98.48  & 96.19 \\
Gemini 2.5 Pro       & 91.00  & 95.83  & 85.59 \\
O3-high              & 94.00  & 96.21  & 91.50 \\
O4-mini-high         & 92.60  & 95.08  & 89.83 \\
O3-mini-high         & 91.40  & 94.32  & 88.14 \\
GPT 4.5              & 89.60  & 91.20  & 87.70 \\
GPT 4.1              & 86.80  & 86.74  & 86.86 \\
GPT 4o               & 85.00  & 83.05  & 86.74 \\
GPT 4o-mini          & 65.40  & 69.07  & 62.12 \\
GPT 4.1 mini         & 77.60  & 75.76  & 79.66 \\
Deepseek R1          & 88.60  & 92.42  & 84.32 \\
Claude 3.7 Sonnet    & 73.60  & 73.11  & 74.15 \\
Llama 3.3            & 70.40  & 69.32  & 71.61 \\
Llama 3              & 66.60  & 64.02  & 69.49 \\
Llama 4 Maverick     & 72.20  & 75.76  & 69.80 \\
Llama 4 Scout        & 69.60  & 66.67  & 72.88 \\
\bottomrule
\end{tabular}
\end{table}

\subsubsection{C.2 MedXpert QA Body System Accuracy} \hfill\\

\begin{figure*}[!h]
  \centering
  \includegraphics[width=\textwidth]{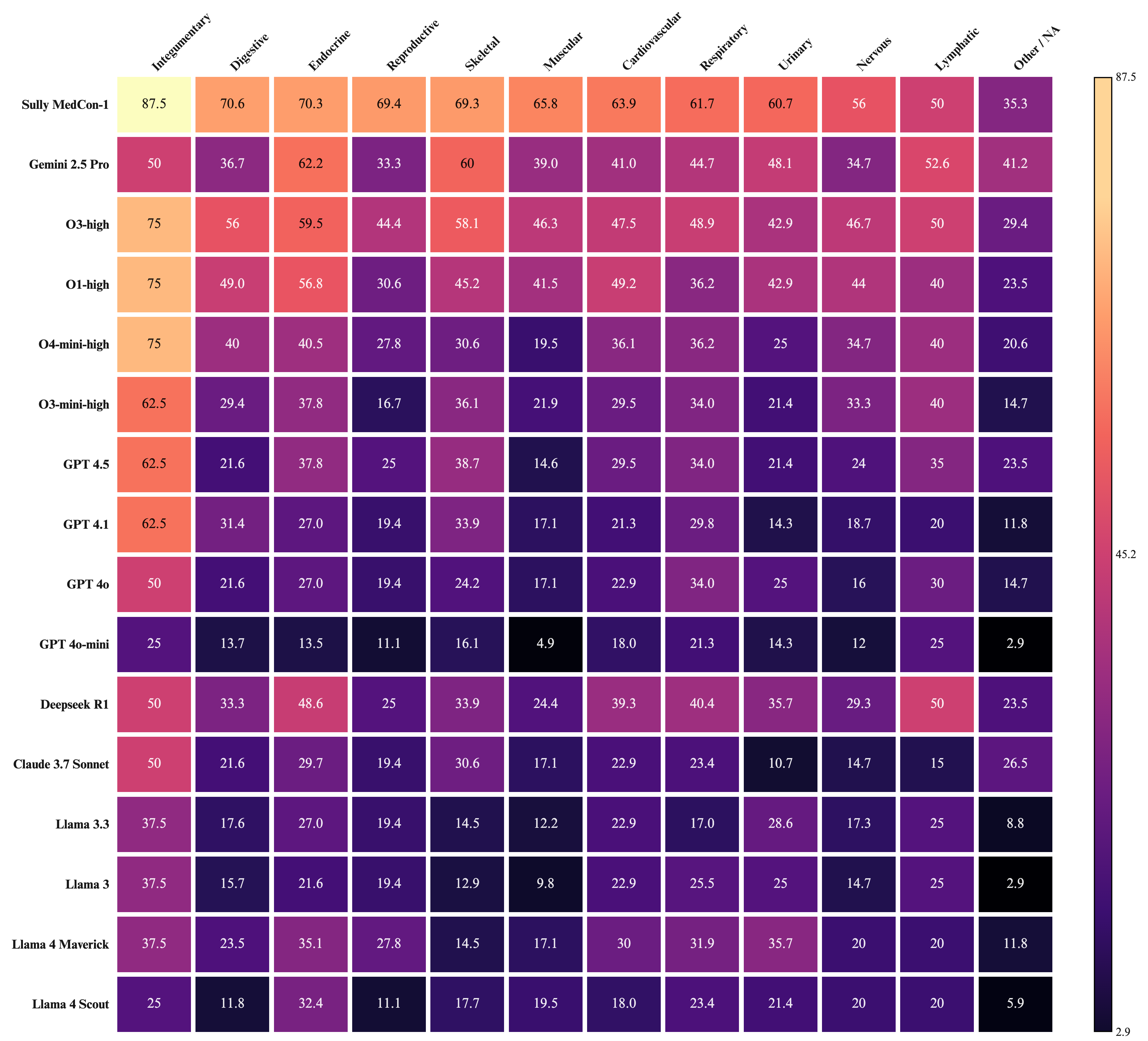}  
  \caption{Body system specific accuracy on the MedXpertQA Evaluation Set}
  \label{fig:fig5}

\end{figure*}
\newpage

\subsubsection{C.3 MedMCQA Specialty Accuracy} \hfill\\
\begin{figure*}[!h]
  \centering
  \includegraphics[width=\textwidth]{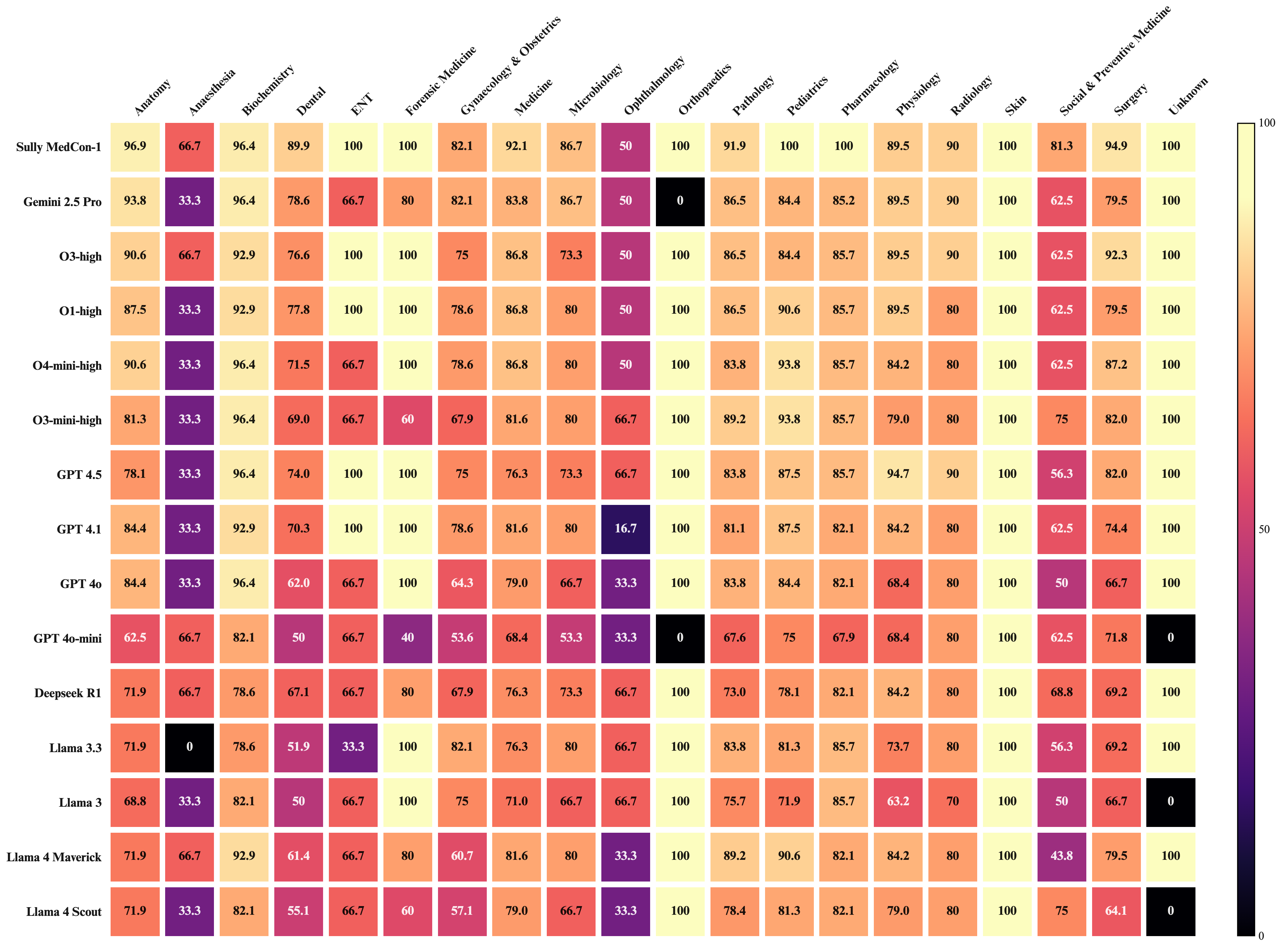}  
  \caption{Medical specialty specific accuracy on MedMCQA Evaluation Set}
  \label{fig:fig5}

\end{figure*}

\newpage

\subsection{Consensus Mechanism Example Output}
\subsubsection{D.0 A Full Consensus Response} \hfill\\

\begin{tcolorbox}[
  title=Task Information,
  colback=green!5,
  colframe=green!40!black,
  fonttitle=\bfseries,
  breakable,
  enhanced jigsaw]

\textbf{Case ID:} Text-101

\textbf{Patient Scenario:} A 56-year-old male presents with severe right leg pain and swelling four days after undergoing a craniotomy for evacuation of a subdural hematoma. He experiences chills and nausea.

\textbf{Medical History:} Type 2 diabetes mellitus, chronic kidney disease requiring hemodialysis for 2 years.

\textbf{Pre-admission Medications:} Insulin, enalapril, atorvastatin, sevelamer.

\vspace{0.5em}
\textbf{Vitals:}
\begin{itemize}[leftmargin=2em]
  \item Temperature: 38.3°C (101.0°F)
  \item Pulse: 110/min
  \item Blood pressure: 130/80 mm Hg
\end{itemize}

\textbf{Physical Exam:} Right calf is swollen, warm, erythematous; pain with dorsiflexion. Peripheral pulses palpable. Cardiopulmonary exam unremarkable.

\vspace{0.5em}
\textbf{Laboratory Results:}
\begin{itemize}[leftmargin=2em]
  \item Hemoglobin: 10.1 g/dL
  \item Leukocyte count: 11,800/mm\textsuperscript{3}
  \item Platelet count: 230,000/mm\textsuperscript{3}
  \item Glucose: 87 mg/dL
  \item Creatinine: 1.9 mg/dL
\end{itemize}

\vspace{0.5em}
\textbf{Question:} What is the most appropriate next step in managing this patient?

\vspace{0.5em}
\textbf{Answer Choices:}
\begin{itemize}[leftmargin=2em]
  \item (A) Surgical thrombectomy
  \item (B) Inferior vena cava filter
  \item (C) Unfractionated heparin therapy
  \item (D) Direct oral anticoagulants (DOACs)
  \item (E) Aspirin therapy only
  \item (F) Prophylactic antibiotics for cellulitis
  \item (G) Oral warfarin therapy
  \item (H) Compression stockings and observation
  \item (I) Thrombolytic therapy with tPA
  \item (J) Low molecular weight heparin therapy
\end{itemize}

\vspace{0.5em}
\textbf{Correct Answer:} (B) Inferior vena cava filter

\vspace{0.5em}
\textbf{Metadata:}
\begin{itemize}[leftmargin=2em]
  \item Task Type: Treatment
  \item Body System: Cardiovascular
  \item Question Type: Reasoning
\end{itemize}

\end{tcolorbox}

\newpage

\begin{tcolorbox}[
  title=Consensus Summary and Final Recommendation,
  colback=blue!5,
  colframe=blue!75!black,
  fonttitle=\bfseries,
  breakable
]
\textbf{Case Summary:} A 56-year-old male presents four days after craniotomy with signs of acute proximal DVT.

\textbf{Key Considerations:}
\begin{itemize}[leftmargin=2em]
  \item Neurosurgical guidelines classify systemic anticoagulation within 7–14 days post-craniotomy as a major contraindication.
  \item LMWH is contraindicated due to both neurosurgical timing and ESRD.
  \item UFH is reversible but still systemic anticoagulation, with caution.
  \item An IVC filter is the standard temporizing measure for proximal DVT when anticoagulation is contraindicated.
\end{itemize}

\textbf{Final Consensus:} Placement of an \textbf{Inferior Vena Cava (IVC) Filter} (Answer: \textbf{B}) as the most appropriate next step.

\textbf{Residual Probability:}
\begin{itemize}
  \item UFH (C): ~30\% – possible with pristine imaging and neurosurgical approval.
  \item LMWH (J): <10\% – contraindicated due to renal clearance and bleed risk.
\end{itemize}
\end{tcolorbox}

\begin{tcolorbox}[
  title=Traige Model Response (Expert Configuration),
  colback=gray!5,
  colframe=black,
  fonttitle=\bfseries,
  coltitle=white,
  breakable
]
\begin{itemize}[leftmargin=2em]
  \item \textbf{Expert 1:} Hematologist
  \item \textbf{Expert 2:} Nephrologist
  \item \textbf{Expert 3:} Vascular Surgeon
  \item \textbf{Expert 4:} Infectious Disease Specialist
\end{itemize}
\end{tcolorbox}

\vspace{1em}

\begin{tcolorbox}[
  title=Expert 1 – Hematologist,
  colback=white,
  colframe=black,
  fonttitle=\bfseries,
  breakable,
  enhanced jigsaw
]

\textbf{Clinical Reasoning:}  
\begin{itemize}[leftmargin=1.5em]
  \item 56-year-old man, POD-4 after craniotomy for subdural hematoma evacuation.
  \item New, unilateral swollen, warm, tender calf with pain on passive dorsiflexion (Homan's sign).
  \item Low-grade fever and mild leukocytosis – common with acute DVT.
  \item Comorbid diabetes, CKD, and dialysis suggest a pro-thrombotic state.
  \item High pre-test probability for acute proximal DVT.
\end{itemize}

\vspace{0.5em}
\textbf{Hematologic/Neurosurgical Risk Assessment:}
\begin{itemize}[leftmargin=1.5em]
  \item Recent intracranial surgery is a major contraindication to full-dose anticoagulation (ACCP, ASH, SVS, NCC guidelines).
  \item Re-bleed risk post-craniotomy is highest in the first 7–14 days; studies show 2–6\% symptomatic re-bleed with early heparin use.
  \item IVC filter recommended when anticoagulation is temporarily contraindicated.
\end{itemize}

\vspace{0.5em}
\textbf{Scientific/Pathophysiologic Reasoning:}
\begin{itemize}[leftmargin=1.5em]
  \item Risk of PE from thrombus propagation.
  \item Anticoagulants (UFH, LMWH, DOACs) impair hemostasis and increase bleed risk.
  \item IVC filter intercepts emboli without systemic anticoagulation.
  \item LMWH contraindicated in ESRD; DOACs not recommended at CrCl < 15 mL/min.
\end{itemize}

\vspace{0.5em}
\textbf{Elimination of Answer Choices:}
\begin{itemize}[leftmargin=1.5em]
  \item A – Surgical thrombectomy: Reserved for limb-threatening clots.
  \item B – IVC filter: Recommended in this case due to absolute contraindication to anticoagulation.
  \item C – UFH: Reversible but still risky in first 7–14 days post-craniotomy.
  \item D – DOACs: Contraindicated in both neurosurgery and ESRD.
  \item E – Aspirin: Ineffective for acute DVT.
  \item F – Antibiotics: No clear signs of cellulitis; risks missing a life-threatening VTE.
  \item G – Warfarin: Slow onset; requires heparin bridge, which is contraindicated.
  \item H – Compression/observation: Does not protect from PE.
  \item I – Systemic tPA: Contraindicated post-craniotomy.
  \item J – LMWH: Renally cleared, long-acting, high risk of accumulation in ESRD.
\end{itemize}

\vspace{0.5em}
\textbf{Final Recommendation:}  
\textbf{B – Inferior vena cava filter} is the most appropriate next step due to immediate protection without increasing intracranial bleeding risk.

\vspace{0.25em}
\textbf{Second Most Likely Answer:}  
C – Unfractionated heparin, if neurosurgical team confirms stable operative field and PE risk is deemed higher than ICH risk.

\vspace{0.75em}
\textbf{Mind Map of Reasoning:}

\begin{tcolorbox}[colback=gray!10, colframe=gray!50, fontupper=\ttfamily, left=1mm, right=1mm, boxrule=0.3pt]
DVT clues (swelling, pain, fever)              Recent craniotomy (Day 4)
       |                                          |
       |                                          |
  ---> ACUTE PROXIMAL DVT  <----------------------+
             |
             |  Dialysis → DOAC/LMWH less suitable
             |  Risk of PE → needs protection
             |
    Management decision tree
    ├── Anticoagulation   (heparin/LMWH/DOAC) – NO (bleed risk)
    ├── Thrombolysis – NO (ICH risk extreme)
    ├── Surgical thrombectomy – Only limb-threatening
    └── Mechanical protection → IVC FILTER (Yes)
\end{tcolorbox}

\vspace{0.5em}
\textbf{Unknowns / Additional Considerations:}
\begin{itemize}[leftmargin=1.5em]
  \item Awaiting duplex ultrasound for confirmation.
  \item Future coordination with neurosurgery for anticoagulation timing.
  \item Consider filter retrieval when safe; continue mechanical calf pumps.
\end{itemize}

\end{tcolorbox}

\begin{tcolorbox}[
  title=Expert 2 – Nephrologist,
  colback=white,
  colframe=black,
  fonttitle=\bfseries,
  breakable,
  enhanced jigsaw
]

\textbf{Clinical Summary:}  
56-year-old male with type 2 diabetes mellitus and end-stage renal disease on hemodialysis presents four days post-craniotomy with right calf pain, swelling, warmth, erythema, and a positive Homan's sign — all highly suggestive of acute DVT. 

\textbf{Major Considerations:}
\begin{itemize}[leftmargin=1.5em]
  \item \textbf{Recent neurosurgery} → significantly increases risk of catastrophic bleeding if anticoagulated.
  \item \textbf{ESRD on dialysis} → renally cleared agents (e.g., LMWH, DOACs) pose risk of accumulation and bleeding.
  \item \textbf{Unfractionated heparin (UFH)} is preferred for its reversibility, short half-life, and non-renal clearance.
\end{itemize}

\vspace{0.5em}
\textbf{Answer Choice Evaluation:}
\begin{itemize}[leftmargin=1.5em]
  \item (A) \textbf{Surgical thrombectomy:} Reserved for limb-threatening cases; not first-line for typical DVT.
  \item (B) \textbf{IVC filter:} Appropriate if anticoagulation is contraindicated. In this case, UFH is deemed safe enough to try first.
  \item (C) \textbf{Unfractionated heparin:} Most appropriate initial treatment; reversible and safe in ESRD.
  \item (D) \textbf{DOACs:} Contraindicated in ESRD; less easily reversed in emergency.
  \item (E) \textbf{Aspirin:} Inadequate for DVT treatment.
  \item (F) \textbf{Antibiotics:} Inappropriate without clear evidence of cellulitis; DVT more likely.
  \item (G) \textbf{Warfarin:} Requires bridging with heparin; too slow as sole therapy.
  \item (H) \textbf{Observation/compression:} Insufficient for confirmed DVT with high PE risk.
  \item (I) \textbf{tPA:} Contraindicated due to recent craniotomy and high ICH risk.
  \item (J) \textbf{LMWH:} Renally cleared; dangerous in ESRD and post-neurosurgical context.
\end{itemize}

\vspace{0.5em}
\textbf{Mind Map of Reasoning:}

\begin{tcolorbox}[colback=gray!10, colframe=gray!50, fontupper=\ttfamily, left=1mm, right=1mm, boxrule=0.3pt]
• Patient Profile: 56-year-old male, post-craniotomy, DM, CKD on dialysis  
    ├─ Signs/Symptoms: Right calf swelling, warmth, erythema, pain on dorsiflexion  
    ├─ Laboratory: Mild leukocytosis, creatinine elevated  
    ├─ Differential: DVT vs. cellulitis (DVT favored by exam)

• Treatment Considerations:  
    ├─ Need for rapid anticoagulation  
    ├─ Recent neurosurgery → bleeding risk → prefer reversible anticoagulants  
    ├─ CKD on dialysis → avoid LMWH and DOACs  

• Answer Evaluation:  
    ├─ (C) UFH: Reversible, short half-life, OK in CKD  
    ├─ (B) IVC filter: Backup if anticoagulation not possible  
    └─ Others: Do not meet safety/efficacy needs  
\end{tcolorbox}

\vspace{0.5em}
\textbf{Final Recommendation:}  
\textbf{C – Unfractionated heparin} is the most appropriate next step, balancing safety in renal impairment and reversibility after recent surgery.

\vspace{0.25em}
\textbf{Second Most Likely Answer:}  
\textbf{B – IVC filter} would be appropriate if UFH were contraindicated due to bleeding risk.

\end{tcolorbox}

\begin{tcolorbox}[
  title=Expert 3 – Vascular Surgeon,
  colback=white,
  colframe=black,
  fonttitle=\bfseries,
  breakable,
  enhanced jigsaw
]

\textbf{Clinical Summary:}  
The patient presents with classic signs of deep vein thrombosis (DVT) — leg swelling, warmth, erythema, pain on dorsiflexion — following a recent craniotomy. The clinical suspicion is further supported by leukocytosis and his known history of chronic kidney disease requiring hemodialysis.

\textbf{Reasoning and Recommendation:}
\begin{itemize}[leftmargin=1.5em]
  \item The post-operative state and CKD create a pro-thrombotic environment.
  \item Anticoagulation is considered the cornerstone of treatment for confirmed or highly suspected DVT.
  \item \textbf{LMWH is preferred} due to its reported efficacy and perceived safety in renal impairment.
  \item \textbf{UFH is offered as a close alternative} if LMWH is contraindicated or if close titration and reversibility are required.
\end{itemize}

\vspace{0.5em}
\textbf{Final Recommendation:}  
\textbf{J – Low molecular weight heparin (LMWH)} is recommended as the initial treatment.

\vspace{0.25em}
\textbf{Second Most Likely Answer:}  
\textbf{C – Unfractionated heparin (UFH)}, particularly if monitoring is required or renal clearance is a concern.

\end{tcolorbox}

\begin{tcolorbox}[
  title=Expert 4 – Infectious Disease Specialist,
  colback=white,
  colframe=black,
  fonttitle=\bfseries,
  breakable,
  enhanced jigsaw
]

\textbf{Clinical Presentation and Differential Diagnosis:}  
\begin{itemize}[leftmargin=1.5em]
  \item 56-year-old male, 4 days post-craniotomy for subdural hematoma evacuation.
  \item Right calf swelling, warmth, erythema, pain on dorsiflexion (Homan's sign).
  \item Low-grade fever (38.3 °C), leukocytosis (11,800/mm³).
  \item High suspicion for DVT given pro-thrombotic context (DM, ESRD, immobility).
  \item Differential includes cellulitis and superficial thrombophlebitis, though DVT is favored.
  \item Imaging (duplex ultrasound) assumed positive or pending.
\end{itemize}

\vspace{0.5em}
\textbf{Key Clinical/Subtle Features:}
\begin{itemize}[leftmargin=1.5em]
  \item Recent intracranial surgery creates high risk for hemorrhage with anticoagulation.
  \item ESRD alters pharmacokinetics — LMWH and DOACs pose renal clearance risks.
  \item Inflammatory signs (fever, leukocytosis) likely due to DVT, not infection.
\end{itemize}

\vspace{0.5em}
\textbf{Answer Choice Evaluation (with Probabilities):}
\begin{itemize}[leftmargin=1.5em]
  \item A – Surgical thrombectomy: Rarely indicated; only for limb-threat (0.02)
  \item B – \textbf{IVC filter:} Appropriate when anticoagulation is contraindicated post-craniotomy (0.55)
  \item C – UFH: Reversible, short half-life, but still poses early bleed risk (0.25)
  \item D – DOACs: Contraindicated in ESRD and early post-craniotomy (0.03)
  \item E – Aspirin: Inadequate for treatment (0.01)
  \item F – Antibiotics: No strong evidence of cellulitis (0.01)
  \item G – Warfarin: Slow onset, needs heparin bridging (0.03)
  \item H – Observation/compression: Not sufficient for proximal DVT (0.02)
  \item I – tPA: Strongly contraindicated post-neurosurgery (0.02)
  \item J – LMWH: Poor choice due to renal clearance and bleeding risk (0.06)
\end{itemize}

\vspace{0.5em}
\textbf{Final Recommendation:}  
\textbf{B – Inferior vena cava filter} is the best option due to high bleed risk and immediate need for PE prevention.

\vspace{0.25em}
\textbf{Second Most Likely Answer:}  
\textbf{C – Unfractionated heparin}, preferred if absolutely necessary, due to its reversibility.

\vspace{0.75em}
\textbf{Mind Map of Reasoning:}

\begin{tcolorbox}[colback=gray!10, colframe=gray!50, fontupper=\ttfamily, left=1mm, right=1mm, boxrule=0.3pt]
1) Patient Profile:
   ├─ Age: 56
   ├─ Comorbidities: DM2, CKD on HD
   └─ Recent Event: Craniotomy for SDH (Day 4)

2) Presentation:
   ├─ Right calf swelling, erythema, warmth
   ├─ Dorsiflexion pain (Homan's sign)
   ├─ Fever, leukocytosis
   └─ No cardiopulmonary findings

3) Differential:
   ├─ DVT
   ├─ Cellulitis
   └─ Superficial thrombophlebitis

4) Diagnosis: Presumed proximal DVT (duplex US pending/high clinical suspicion)

5) Management Constraints:
   ├─ High bleeding risk (post-craniotomy)
   ├─ CKD: LMWH/DOAC accumulation
   └─ Need to prevent PE

6) Treatment Options:
   ├─ Anticoagulation (UFH, LMWH, DOACs, warfarin)
   ├─ IVC filter
   ├─ Thrombolysis
   ├─ Surgical options
   └─ Conservative care

7) Best Choice: IVC filter (prevents PE, no systemic bleed risk)
8) Second Best: UFH (reversible, titratable)

Unknowns Considered:
- Duplex ultrasound confirmation
- Extent of thrombus (proximal vs distal)
- Neurological stability and neurosurgical input
- Hematology consultation if needed
\end{tcolorbox}

\end{tcolorbox}

\end{document}